\documentclass{article}

\usepackage{amsmath,amsthm,amssymb,amsfonts}

\usepackage[a4paper]{geometry}

\usepackage{a4wide}

\usepackage{graphicx}

\usepackage{subfigure}

\usepackage{url}

\usepackage{hyperref}

\usepackage{listings}
\usepackage{color}
\usepackage[dvipsnames]{xcolor}

\usepackage[margin=10pt,font=small,labelfont=bf, labelsep=endash]{caption}

\usepackage{bbm}
\usepackage{algorithm}
\usepackage{algpseudocode}

\numberwithin{theorem}{section}

\theoremstyle{definition}

\numberwithin{equation}{section}

        \def\l{\lambda }    \def\r{\rho}

\newcommand{\sF}{{\cal F}}
\newcommand{\sG}{{\cal G}}

\newcommand{\sP}{{\cal P}}

\newcommand{\R}{{\mathbb R}}

\newcommand{\rottext}[1]{\rotatebox{90}{\parbox{24mm}{\small \centering#1}}}

\newcommand{\lt}{\left}
\newcommand{\rt}{\right}

\usepackage{environ}
\newcommand{\acksection}{\section*{Acknowledgments and Disclosure of Funding}}
\NewEnviron{ack}{%
	\acksection
	\BODY
}

\usepackage{mathtools}

\newcommand{\pa}[1]{\left(#1\right)} % encloses the argument using stretchable parentheses
\newcommand{\mat}[1]{\begin{matrix*}[c] #1\end{matrix*}} % no delimiters
\newcommand{\pmat}[1]{\pa{\mat{#1}}} % parentheses as delimiters
\newcommand{\sm}[1]{\begin{smallmatrix*}[r] #1\end{smallmatrix*}} % no delimiters 
\newcommand{\psm}[1]{\pa{\sm{#1}}} % parentheses as delimiters
\newcommand{\edits}[1]{#1}

\newcommand*\samethanks[1][\value{footnote}]{\footnotemark[#1]}

\title{Alternating the Population and Control Neural Networks to Solve High-Dimensional Stochastic Mean-Field Games}
\author{Alex Tong Lin\thanks{equal contribution}\protect\phantom{\footnotesize 1}\textsuperscript{,}\thanks{University of California, Los Angeles}
	\and Samy Wu Fung\samethanks[1]\protect\phantom{\footnotesize 1}\textsuperscript{,}\samethanks[2]
	\and Wuchen Li \thanks{University of South Carolina}
	\and Levon Nurbekyan \samethanks[2] 
	\and Stanley J. Osher \samethanks[2]}
\date{}

\begin{document}
	
	\maketitle
	
	\begin{abstract}
		We present APAC-Net, an alternating population and agent control neural network for solving stochastic mean field games (MFGs). Our algorithm is geared toward high-dimensional instances of MFGs that are beyond reach with existing solution methods. We achieve this in two steps. First, we take advantage of the underlying variational primal-dual structure that MFGs exhibit and phrase it as a convex-concave saddle point problem. Second, we parameterize the value and density functions by two neural networks, respectively. By phrasing the problem in this manner, solving the MFG can be interpreted as a special case of training a generative adversarial network (GAN). We show the potential of our method on up to 100-dimensional MFG problems.
	\end{abstract}

	\section{Introduction}
	Mean field games (MFGs) are a class of problems that model large populations of interacting agents. They have been widely used in economics~\cite{moll14,moll17,gueant2011mean,gomes2015economic}, finance~\cite{caines17,cardialiaguet2018,jaimungal19,moll17}, industrial engineering~\cite{paola19,kizikale19,gomes2018electricity}, swarm robotics~\cite{liu2018mean,elamvazhuthi2019mean}, epidemic modelling~\cite{lee2020controlling,chang2020game} and data science~\cite{weinan2019mean,guo2019learning,carmona2019linear}. In mean field games, a continuum population of small rational agents play a non-cooperative differential game on a time horizon $[0,T]$. 
	At the optimum, the agents reach a Nash equilibrium, where they can no longer unilaterally improve their objectives. Given the initial distribution of agents $\rho_0 \in \mathcal{P}(\R^n)$, where $\mathcal{P}(\R^n)$ is the space of all probability densities, the solution to MFGs are obtained by solving the system of partial differential equations (PDEs), 
	\begin{equation}
	\label{eq:MFGPDEs}
	\begin{split}
	&-\partial_t \phi - \nu \Delta \phi + H(x,\nabla \phi) = f(x,\rho) \qquad \text{ (HJB)}
	\\
	&\partial_t \rho - \nu \Delta \rho - \text{div} (\rho \nabla_p H(x,\nabla \phi)) = 0 \qquad \text{ (FP)}
	\\
	&\rho(x,0) = \rho_0, \quad \phi(x,T) = g(x,\rho(\cdot,T))
	\end{split}
	\end{equation}
	which couples a Hamilton-Jacobi-Bellman (HJB) equation and a Fokker-Planck (FP) equation.
	Here, $\phi\colon \R^n \times [0,T] \to \R$ is the value function, i.e., the policy that guides the agents, $H\colon \R^n \times \R^n \to \R$ is the Hamiltonian, which describes the physics of the environment,  $\rho(\cdot, t) \in \mathcal{P}(\R^n)$ is the distribution of agents at time $t$, $f\colon \R^n \times \mathcal{P}(\R^n) \to \R$ denotes the interaction between the agents and the population, and $g\colon \R^n \times \mathcal{P}(\R^n) \to \R$ is the terminal condition, which guides the agents to the final distribution.
	Under standard assumptions, i.e. convexity of $H$ in the second variable, and monotonicity \edits{of $f$ and $g$ -- namely that} 
	\begin{equation*}
	\edits{\int_\R^n (f(y, \rho_1) - f(y, \rho_2))d(\rho_1 - \rho_2)(y) > 0 \text{ for all } \rho_1\neq\rho_2, }
	\end{equation*}
	\edits{and similarly for $g$} -- then the solution to~\eqref{eq:MFGPDEs} exists and is unique. See~\cite{lasry2006jeux,LasryLions2007,chow2017algorithm} for more details.
	Although there is a plethora of fast solvers for the solution of~\eqref{eq:MFGPDEs} in two and three dimensions~\cite{achdou2010mean,benamou2017variational,chow2019algorithm,chow2017algorithm,chow2018algorithm,jacobs2019solving}, numerical methods for solving~\eqref{eq:MFGPDEs} in high dimensions are practically nonexistent due to the need for grid-based spatial discretization. These grid-based methods are prone to the curse of dimensionality, i.e., their computational
	complexity grows exponentially with spatial dimension~\cite{bellman1966dynamic}.
	Thus, grid-based methods cannot be tractably used on, e.g., modeling an energy efficient heating, ventilation, and air conditioning system in a complex building, where the dimensions can be as high as  1000~\cite{li2016mean}. 
	
	\paragraph{Our Contribution} We present APAC-Net, an alternating population and agent control neural network approach for tractably solving high-dimensional MFGs in the stochastic case ($\nu > 0$). To this end, we phrase the MFG problem as a saddle-point problem \cite{LasryLions2007,benamou2017variational,nurbekyan18} and parameterize the value function \emph{and} the density function. This formulation allows us to circumvent using spatial grids or uniformly sampling in high dimensions, i.e., the curse of dimensionality.
	While spatial grids for MFGs are also avoided in~\cite{ruthotto2020machine}, their work is limited to the deterministic setting $(\nu=0)$.
	Thus, to the best of our knowledge, APAC-Net is the first model that solves high-dimensional MFGs in the stochastic setting $(\nu>0)$. APAC-Net does this by drawing from a natural connection between MFGs and generative adversarial neural networks (GANs)~\cite{goodfellow2014generative} (see~\ref{sec:connections}), a powerful class of generative models that have shown remarkable success on various types of datasets~\cite{goodfellow2014generative,arjovsky2017wasserstein,gulrajani2017improved,lin2018wasserstein,dukler2019wasserstein,chu2020stability}.
	
	\section{Variational Primal-Dual Formulation of Mean Field Games}\label{sec:mathFormulation}
	
	We derive the mathematical formulation of MFGs for our framework; in particular, we arrive at a primal-dual convex-concave formulation tailored for our alternating networks approach.
	An MFG system \eqref{eq:MFGPDEs} is called potential, if there exist functionals $\sF,\sG$ such that 
	\begin{equation}
	\delta_\rho \sF=f(x,\rho) \quad \text{ and } \quad \delta_\rho \sG=g(x,\rho),
	\end{equation}
	where
	\begin{equation} \label{eq:potentialDef}
	\begin{split}
	&\langle \delta_\rho \sF(\rho),\mu \rangle=\lim\limits_{h\to 0} \frac{\sF(\rho+h \mu)-\sF(\rho)}{h},~\forall~\mu, \\
	&\langle \delta_\rho \sG(\rho),\mu \rangle=\lim\limits_{h\to 0} \frac{\sG(\rho+h \mu)-\sG(\rho)}{h},~\forall~\mu.
	\end{split}
	\end{equation}
	That is, there exist functionals $\mathcal{F}, \mathcal{G}$ such that their variational derivatives with respect to $\rho$ are the interaction and terminal costs $f$ and $g$ from~\eqref{eq:MFGPDEs}.
	A critical feature of potential MFGs is that the solution to~\eqref{eq:MFGPDEs} can be formulated as the solution to a convex-concave saddle point optimization problem. 
	To this end, we begin by stating~\eqref{eq:MFGPDEs} as a variational problem~\cite{LasryLions2007,benamou2017variational} akin to the Benamou-Brenier formulation for the Optimal Transport (OT) problem:
	\begin{equation}\label{eq:originalMFGVariationaProb}
	\begin{split}
	&\inf_{\rho, v}  \int_0^T \left\{\int_\Omega  \rho(x,t) L(x, v(x,t))dx\, + \sF(\rho(\cdot,t))\right\}   dt + \sG(\rho(\cdot,T))
	\\
	&\text{ s.t. } \partial_t \rho - \nu \Delta \rho + \nabla \cdot (\rho v) = 0, \quad \rho(x,0) = \rho_0(x),
	\end{split}
	\end{equation}
	where $L\colon \R^n \times \R^n \to \R$ is the Lagrangian function corresponding to the Legendre transform of the Hamiltonian $H$, and $\sF,\sG \colon \R^n \times \sP(\R^n) \to \R$ are mean field interaction terms, and $v \colon \R^n \times [0,T] \to \R^n$ is the velocity field. Next, setting $\phi$ as a Lagrange multiplier, we insert the PDE constraint into the objective to get
	\begin{equation}
	\begin{split}
	\sup_{\phi} \inf_{\rho(x,0)=\rho_0(x), v} &\int_0^T \left\{\int_\Omega  \rho(x,t) L(x, v(x,t))dx\, + \sF(\rho(\cdot,t))\right\} dt + \sG(\rho(\cdot,T)) \\
	&- \int_0^T \int_\Omega \phi(x,t)\lt(\partial_t \rho - \nu \Delta \rho + \nabla \cdot(\rho(x,t) v(x,t) \rt)\,dx \,dt.
	\end{split}
	\end{equation}
	Finally, integrating by parts and minimizing with respect to $v$ to obtain the Hamiltonian via 
 %$H(x,p)=\inf_v \left\{-p\cdot v+L(x,v)\right\}$,
    $-H(x,-p) = \inf_v \left\{ L(x,v) - p\cdot v\right\}$
    we obtain
	\begin{equation}\label{eq:minmax_MFG_gen}
	\begin{split}
	\inf_{\rho(x,0)=\rho_0(x)} \sup_{\phi} &\int_0^T \left\{\int_\Omega \lt(\partial_t \phi+\nu \Delta \phi - H(x,\nabla \phi)\rt) \rho(x,t)\,dx  + \sF(\rho(\cdot,t))\,\right\}dt\\
	&+\int_\Omega \phi(x,0)\rho_0(x)dx-\int_{\Omega}\phi(x,T)\rho(x,T)dx+\sG(\rho(\cdot,T)).
	\end{split}
	\end{equation}
	We note that in our numerical experiments, we will actually have $\sG(\rho(\cdot,T)) = \int_\Omega g(x) \rho(x, T)\, dx$, with $g(x)$ the terminal condition of $\phi$.
    Here our approach follows that of  \cite{benamou2017variational, cardaliaguet2015mean, cardaliaguet2015second}. This formula can also be obtained in the context of HJB equations in density spaces \cite{chow2019algorithm}, or by integrating the HJB and the FP equations in \eqref{eq:MFGPDEs} with respect to $\rho$ and $\phi$, respectively \cite{nurbekyan18}. 
	In \cite{nurbekyan18}, it was observed that all MFG systems admit an infinite-dimensional two-player general-sum game formulation, and the potential MFGs are the ones that correspond to zero-sum games.
	In this interpretation, Player 1 represents the \emph{mean-field} or the \emph{population as a whole} and their strategy is the population density $\rho$. Furthermore, Player 2 represents the \textit{generic agent} and their strategy is the value function $\phi$. The aim of Player 2 is to provide a strategy that yields the best response of a generic agent against the population. This interpretation is in accord with the intuition behind generative adversarial networks (GANs), as the key observation is that under mild assumptions on $\mathcal{F}$ and $\mathcal{G}$, each spatial integral is really an expectation from $\rho$. 
	The formulation \eqref{eq:minmax_MFG_gen} is the cornerstone of our method.
	
	\section{Connections to GANs}\label{sec:connections}
	\paragraph{Generative Adversarial Networks}
	
	In GANs \cite{goodfellow2014generative}, we have a discriminator and generator, and the goal is to obtain a generator that is able to produce samples from a desired distribution. 
	The generator does this by taking samples from a known distribution $\mathcal{N}$ and transforming them into samples from the desired distribution. 
	Meanwhile, the purpose of the discriminator is to aid the optimization of the generator. Given a generator network $G_\theta$ and a discriminator network $D_\omega$, the original GAN objective is to find an equilibrium to the minimax problem
	\begin{equation}
	\inf_{G_\theta} \sup_{D_\omega} \mathbb{E}_{x\sim \rho_0}\lt[ \log D_\omega(x)\rt] + \mathbb{E}_{z\sim \mathcal{N}}\lt[ \log(1 - D_\omega(G_\theta(z))) \rt]. 
	\end{equation}
	Here, the discriminator acts as a classifier that attempts to distinguish real images from fake/generated images, and the goal of the generator is to produce samples that ``fool" the discriminator.
	
	\paragraph{Wasserstein GANs}
	
	In Wasserstein GANs \cite{arjovsky2017wasserstein}, the motivation is drawn from OT theory, where now the objective function is changed to the Wasserstein-1 (W1) distance in the Kantorovich-Rubenstein dual formulation
	\begin{equation}
	\inf_{G_\theta} \sup_{D_\omega} \mathbb{E}_{x\sim \rho_0} \lt[D_\omega(x)\rt] - \mathbb{E}_{z\sim \mathcal{N}} \lt[ D_\omega(G_\theta(z))\rt], \text{ s.t. }  \edits{\|\nabla D_\omega\| \le 1},
	\end{equation}
	and the discriminator is required to be 1-Lipschitz. In this setting, the goal of the discriminator is to compute the W1 distance between the distribution of $\rho_0$ and $G_\theta(z)$. In practice, using the W1 distance helps prevent the generator from suffering "mode collapse," a situation where the generator produces samples from only one mode of the distribution $\rho_0$; for instance, if $\rho_0$ is the distribution of images of handwritten digits, then mode collapse entails producing only, say, the 0 digit. Originally, weight-clipping was to enforce the Lipschitz condition of the discriminator network~\cite{arjovsky2017wasserstein}, but an improved method using a penalty on the gradient was used in \cite{gulrajani2017improved}.
	
	\paragraph{GANs $\leftrightarrow$ MFGs}
	A Wasserstein GAN can be seen as a particular instance of a deterministic MFG~\cite{cao2020connecting,benamou2017variational,LasryLions2007}. 
	Specifically, consider the MFG~\eqref{eq:minmax_MFG_gen} in the following setting. 
	Let $\nu = 0$, $\mathcal{G}$ be a hard constraint with target measure $\rho_T$ (as in optimal transport), and let $H$ be the Hamiltonian defined by
	\begin{equation}
	H(x,p) = \mathbbm{1}_{\|p\|\le 1} = 
	\begin{cases} 
	0 & \| p \|\leq 1 \\
	\infty & \text{otherwise}
	\end{cases},
	\end{equation}
	where we note that this Hamiltonian arises when the Lagrangian is given by $L(x,v) = \|v \|_2$.
	Then \eqref{eq:minmax_MFG_gen} reduces to,
	\begin{equation}
	\begin{split}
	&\sup_{\phi} \int_\Omega \phi(x)\rho_0(x)\, dx - \int_{\Omega}\phi(x)\rho_T(x)\, dx\\
	&\qquad \text{s.t. } \|\nabla \phi(x)\| \le 1, 
	\end{split}
	\end{equation}
	where we note that the optimization in $\rho$ leads to $\partial_t \phi-H(x,\nabla \phi)=0$. And since $H(p) = \mathbbm{1}_{\|p\|\le 1}$, we have that $\partial_t \phi = 0$, and $\phi(x,t) = \phi(x)$ for all $t$. 
	We observe that the above is precisely the Wasserstein-1 distance in the Kantorovich-Rubenstein duality~\cite{Villani2003}.
	
	\section{APAC-Net}\label{sec:APAC-Net}
	Rather than discretizing the domain and solving for the function values at grid-points, APAC-Net avoids them by parameterizing the function and solving for the function itself. We make a small commentary that this rhymes with history, in moving from the Riemann to the Lebesgue integral: the former focused on grid-points of the domain, whereas the latter focused on the function values.
	
	The training process for our MFG is similar to that of GANs. We initialize the neural networks $N_\omega(x,t)$ and $N_\theta(z,t)$. We then let
	\begin{equation}
	\label{eq:NNStructure}
	\phi_\omega(x, t) = (1 - t) N_\omega(x, t) + t g(x), \; G_\theta(z, t) = (1 - t)z + t N_\theta(z, t),
	\end{equation}
	where $z \sim \rho_0$ are samples drawn from the initial distribution. \edits{Thus, we set $\rho(\cdot, t) = G_\theta(\cdot, t)\#\rho_0$, i.e., the pushforward of $\rho_0$. In this setting,
		we train $G_\theta(\cdot, t)$ to produce samples from $\rho(\cdot,t)$.}
	We note that $\phi_\omega$ and $G_\theta$ automatically satify the terminal and initial condition, respectively. In particular, $G_\theta$ produces samples from $\rho_0$ at $t=0$.
	
	Our strategy for training consists of alternately training $G_\theta$ (the population), and $\phi_\omega$ (the value function for an individual agent). 
	Intuitively, this means we are \emph{alternating the population and agent control neural networks} (APAC-Net) in order to find the equilibrium. 
	Specifically, we train $\phi_\omega$ by first sampling a batch $\{z_b\}_{b=1}^B$ from the given initial density $\rho_0$, and $\{t_b\}_{b=1}^B$ uniformly from $[0,1]$; so we are really sampling from the products of the densities $\rho_0$ and $\text{Unif}([0,1])$. Next, we compute the push-forward $x_b = G_\theta(z_b, t_b)$ for $b=1,\ldots, B$. We then compute the loss,
	\begin{equation}
	\begin{split}
	\text{loss}_{\phi} = &\frac{1}{B} \sum_{b=1}^B \phi_\omega(x_b, 0) + \frac{1}{B}\sum_{b=1}^B \partial_t \phi_\omega(x_b, t_b) + \nu \Delta \phi_\omega(x_b, t_b) - H(\nabla_x \phi_\omega(x_b, t_b))
	\end{split}
	\end{equation}
	where we can optionally add a regularization term 
	\begin{equation}
	\begin{split}
	\text{penalty} = \lambda \frac{1}{B}\sum_{b=1}^B &\lt\|\partial_t \phi_\omega(x_b, t_b) + \nu \Delta \phi_\omega(x_b, t_b) - H(\nabla_x \phi_\omega(x_b, t_b)) + f(x_b, t_b)\rt\|
	\end{split}
	\end{equation} 
	to penalize deviations from the HJB equations~\cite{ruthotto2020machine,onken2020otflow}.
	This extra regularization term has also been found effective in, e.g., Wasserstein GANs \cite{gueant2011mean}, where the norm of the gradient (i.e., the HJB equations) is penalized.
	Finally, we backpropagate the loss to the weights of $\phi_w$. 
	To train the generator, we again sample $\{z_b\}_{b=1}^B$ and $\{t_b\}_{b=1}^B$ as before, and compute
	\begin{equation}
	\begin{split}
	\text{loss}_{G} = &\frac{1}{B}\sum_{b=1}^B \partial_t \phi_\omega(G_\theta(z_b), t_b) + \nu \Delta \phi_\omega(G_\theta(z_b), t_b) - H(\nabla_x \phi_\omega(G_\theta(z_b), t_b)) + f(G_\theta(z_b), t_b).
	\end{split}
	\end{equation}
	We note that because of equation \eqref{eq:NNStructure}, then the term $\int_\Omega (g(x) - \phi(x,T))\rho(x,T)\,dx = 0$ automatically, so that $\rho(x, T)$ does not appear in the optimization $\text{loss}_G$.
    Finally, we backpropagate this loss with respect to the weights of $G_\theta$ (see Alg~\ref{alg:APACnet}).
	
	\begin{algorithm}[t]
		\caption{APAC-Net}
		\label{alg:APACnet}
		\begin{algorithmic}
			\State{\textbf{Require:} $\nu$ diffusion parameter, $G$ terminal cost, $H$ Hamiltonian, $f$ interaction term.} 
			\State{\textbf{Require:} Initialize neural networks $N_\omega$ and  $N_\theta$, batch size $B$} 
			\State{\textbf{Require:} Set $\phi_\omega$ and $G_\theta$ as in~\eqref{eq:NNStructure}}
			\While{not converged}
			\State{\textbf{train $\phi_\omega$}:}
			\State{Sample batch $\{(z_b, t_b)\}_{b=1}^B$ where $z_b\sim \rho_0$ and $t_b\sim \text{Unif}(0,T)$}
			\State{$x_b \gets G_\theta(z_b, t_b)$ for $b=1,\ldots, B$.}
			\State{$\ell_0 \gets \frac{1}{B} \sum_{b=1}^B \phi_\omega(x_b, 0)$}
			\State{$\ell_t \gets \frac{1}{B}\sum_{b=1}^B \partial_t \phi_\omega(x_b, t_b) + \nu \Delta \phi_\omega(x_b, t_b) - H(\nabla_x \phi_\omega(x_b, t_b))$}
			\State{$\ell_{\text{HJB}} \gets \lambda \frac{1}{B}\sum_{b=1}^B \|\partial_t \phi_\omega(x_b, t_b) + \nu \Delta \phi_\omega(x_b, t_b) - H(\nabla_x \phi_\omega(x_b, t_b)) + f(x_b, t_b)\|$}
			\State{Backpropagate the loss $\ell_{\text{total}} = \ell_0 + \ell_t + \ell_{\text{HJB}}$ to $\omega$ weights.}
			\State{}
			\State{\textbf{train $G_\theta$}:}
			\State{Sample batch $\{(z_b, t_b)\}_{b=1}^B$ where $z_b\sim \rho_0$ and $t_b\sim \text{Unif}(0,T)$}
			\State{$\ell_t \gets \frac{1}{B}\sum_{b=1}^B \partial_t \phi_\omega(G_\theta(z_b, t_b), t_b) + \nu\Delta \phi_\omega(G_\theta(z_b, t_b), t_b) \newline \hphantom{m}\hphantom{m}\hphantom{m}\hphantom{m}\hphantom{m}\hphantom{m}\hphantom{m} - H(\nabla_x \phi_\omega(G_\theta(z_b, t_b), t_b)) + f(G_\theta(z_b, t_b), t_b)$}
			\State{Backpropagate the loss $\ell_{\text{total}} = \ell_t$ to $\theta$ weights.}
			\EndWhile
		\end{algorithmic}
	\end{algorithm}
	
	\section{Related Works}
	\paragraph{High-dimensional MFGs and Optimal Control} To the best of our knowledge, the first work to solve MFGs efficiently in high dimensions ($d=100$) was done in~\cite{ruthotto2020machine}. Their work consisted of using Lagrangian coordinates and parameterizing the value function using a neural network. 
	Finally, to estimate the densities, the instantaneous change of variables formula~\cite{chen2018neural}. 
	This combination allowed them to successfully avoid using spatial grids when solving deterministic MFG problems $(\nu = 0)$ with quadratic Hamiltonians. 
	Besides only computing MFGs with $\nu=0$, another limitation is that for non-quadratic Hamiltonians, the instantaneous change of variables formula may lead to high computational costs when estimating the density.
	APAC-Net circumvents this limitation by rephrasing the MFG as a saddle point problem~\eqref{eq:minmax_MFG_gen} and using a GAN-based approach to train two neural networks instead.
	For problems involving high-dimensional optimal control and differential games, spatial grids were also avoided~\cite{chow2017algorithm,chow2018algorithm,chow2019algorithm,lin2018splitting,darbon2016algorithms}. 
	However, these methods are based on generating individual trajectories per agent, and cannot be directly applied to MFGs without spatial discretization of the density, thus limiting their use in high dimensions.

	\paragraph{Reinforcement Learning}W
	Our work bears connections with multi-agent reinforcement learning (RL), where neither the Lagrangian $L$ nor the dynamics (constraint) in~\eqref{eq:originalMFGVariationaProb} are known. Here, a key difference is that multi-agent RL generally considers a finite number of players.
	\cite{wai2018multi} proposes a primal-dual distributed method for multi-agent RL.
	\cite{guo2019learning,guo2020general} propose a Q-Learning approach to solve these multi-agent RL problems.
	\cite{carmona2019linear} studies the convergence of policy gradient methods on mean field reinforcement learning (MFRL) problems, i.e., problems where the agents try instead to learn the control which is socially optimal for the entire population.
	\cite{yang2018deep} uses an inverse reinforcement learning approach to learn the MFG model along with its reward function.
	\cite{Fu2020Actor-Critic} proposes an actor-critic method for finding the Nash equilibrium in linear-quadratic mean field games and establish linear convergence.
	
	\paragraph{Generative Modeling with Optimal Transport}
	There is a class of works that focus on using OT, a class of MFGs, to solve problems arising in data science, and in particular, GANs.
	~\cite{pmlr-v84-genevay18a} presents a tractable method to train large scale generative models using the Sinkhorn distance, which consist of loss functions that interpolate between Wasserstein (OT) distance and Maximum Mean Discrepancy (MMD).
	\cite{salimans2018improving} proposes a mini-batch MMD-based distance to improve training GANs. 
	\cite{sanjabi2018convergence} proposes a class of regularized Wasserstein GAN problems with theoretical guarantees.
	\cite{tanaka2019discriminator} uses a trained discriminator from GANs to further improve the quality of generated samples.
	\cite{lin2019fluid} phrases the adversarial problem as a matching problem in order to avoid solving a minimax problem. Finally,~\cite{peyre2019computational} provides an excellent survey on recent numerical methods for OT and their applications to GANs. 
	
	\paragraph{GAN-based Approach for MFGs}
	Our work is most similar to~\cite{cao2020connecting}, where a connection between MFGs and GANs is also made. However, APAC-Net differs from~\cite{cao2020connecting} in two fundamental ways. First, instead of choosing the value function to be the generator, we set the \emph{density} function as the generator. This choice is motivated by the fact that the generator outputs samples from a desired distribution. It is also aligned with other generative modeling techniques arising in continuous normalizing flows~\cite{finlay2020howtotrain,grathwohl2018ffjord,onken2020discretizeoptimize,onken2020otflow}. Second, rather than setting the generator/discriminator losses as the residual errors of~\eqref{eq:MFGPDEs}, we follow the works of~\cite{nurbekyan18,chow2019algorithm,benamou2017variational,LasryLions2007} and utilize the underlying variational primal-dual structure of MFGs, see~\eqref{eq:minmax_MFG_gen}; this allows us to arrive at the Kantorovich-Rubenstein dual formulation of Wasserstein GANs~\cite{Villani2003}. 
	
	\begin{figure}[t]
		\centering
		\setlength{\tabcolsep}{0.5pt}
		\small
		\begin{tabular}{cccc}
			$\nu = 0$ & $\nu = 0.2$ & $\nu = 0.4$ & $\nu = 0.6$
			\\
			\includegraphics[width=0.2\textwidth]{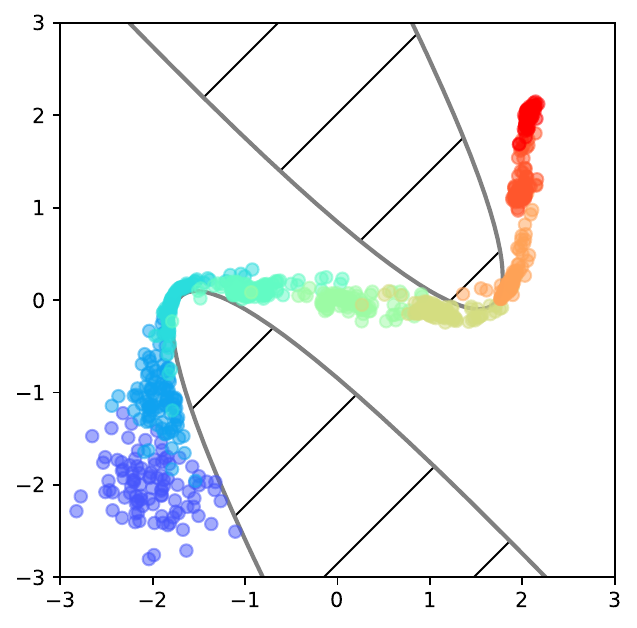}
			&  
			\includegraphics[width=0.2\textwidth]{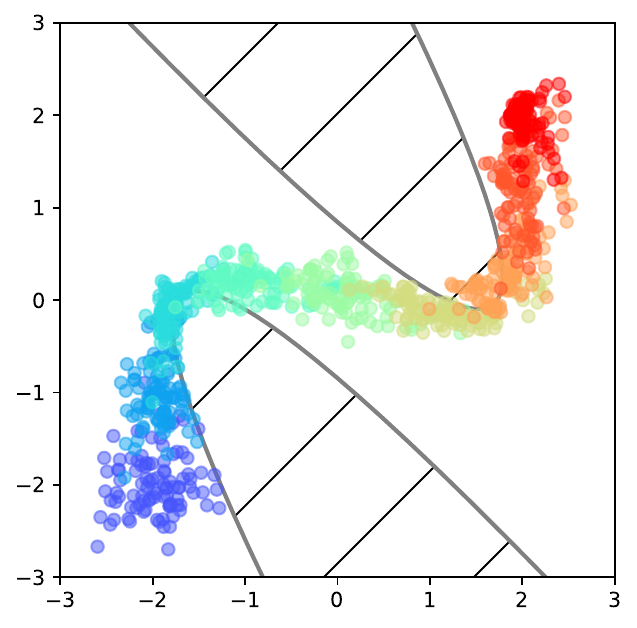}
			& 
			\includegraphics[width=0.2\textwidth]{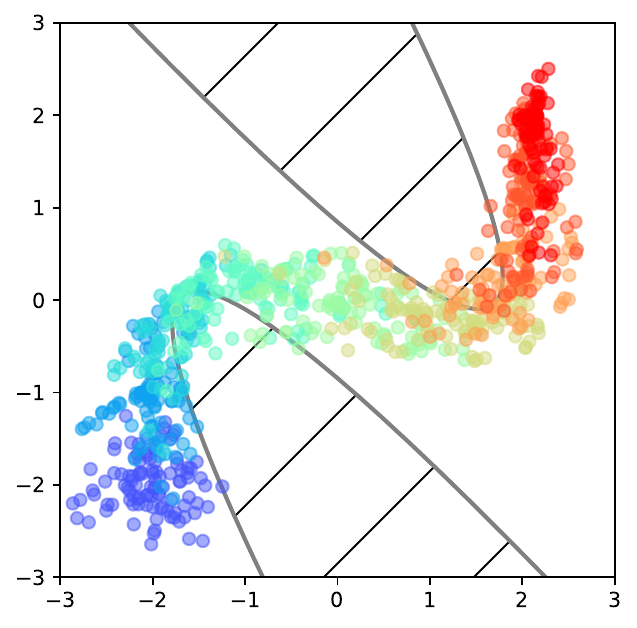}
			& 
			\includegraphics[width=0.2\textwidth]{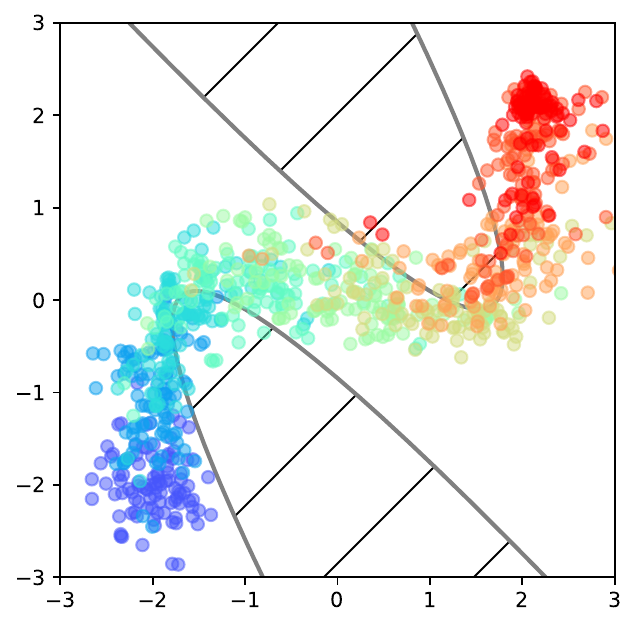}
		\end{tabular}
		\caption{Comparison of 2D solutions for different values of $\nu$. The agents start at the blue points $(t=0)$ and end at the red points $(t=1)$.}
		\label{fig:compareNu}
	\end{figure}
	
	\begin{figure}[t]
		\centering
		\setlength{\tabcolsep}{0.5pt}
		\small
		\begin{tabular}{cccccc}
			\ & $d=2$ & $d=50$ & $d=100$\ \\
			\rottext{$\nu = 0$} & \includegraphics[width=0.2\textwidth]{figures/plot__obstacle__dim-2__nu-0__epoch-200000.pdf}
			&
			\includegraphics[width=0.2\textwidth]{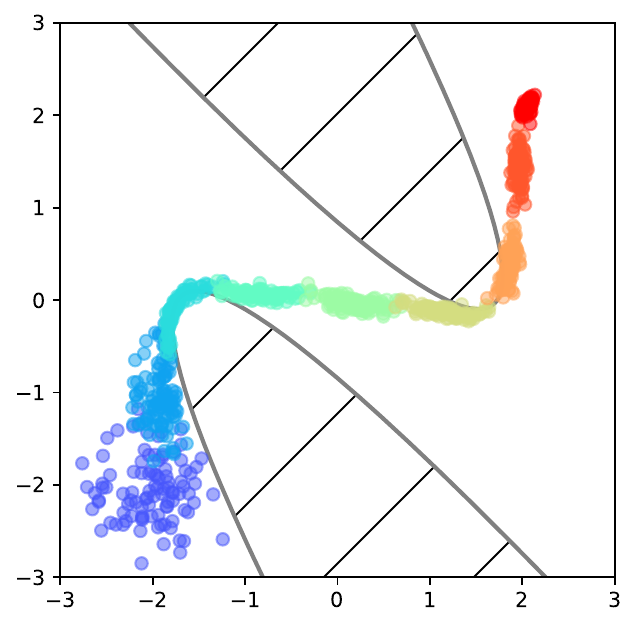}
			&
			\includegraphics[width=0.2\textwidth]{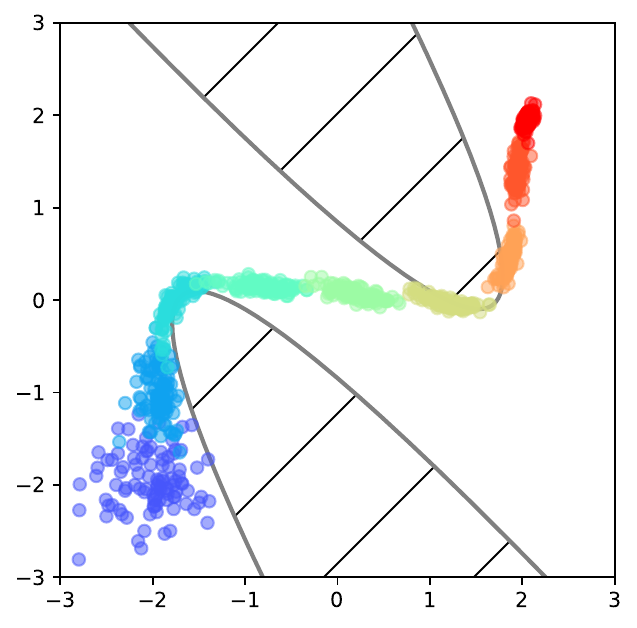}
			\\
			\rottext{$\nu=0.4$} & \includegraphics[width=0.2\textwidth]{figures/plot__obstacle__dim-2__nu-0p4__epoch-200000.pdf}
			& 
			\includegraphics[width=0.2\textwidth]{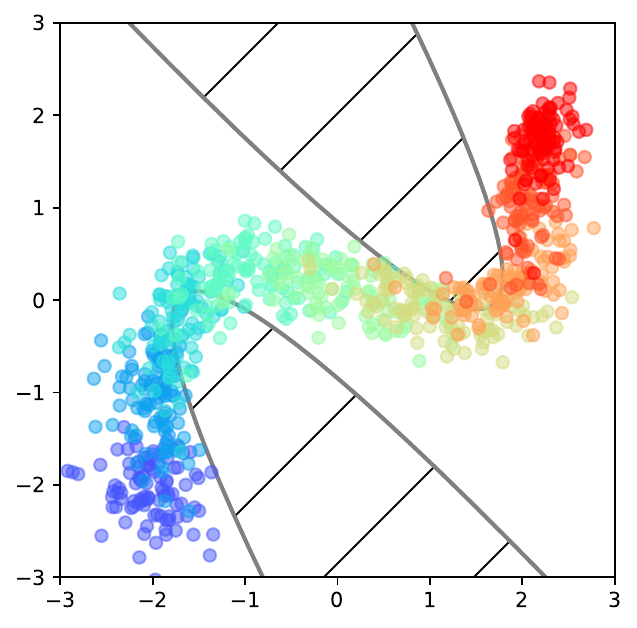}
			& 
			\includegraphics[width=0.2\textwidth]{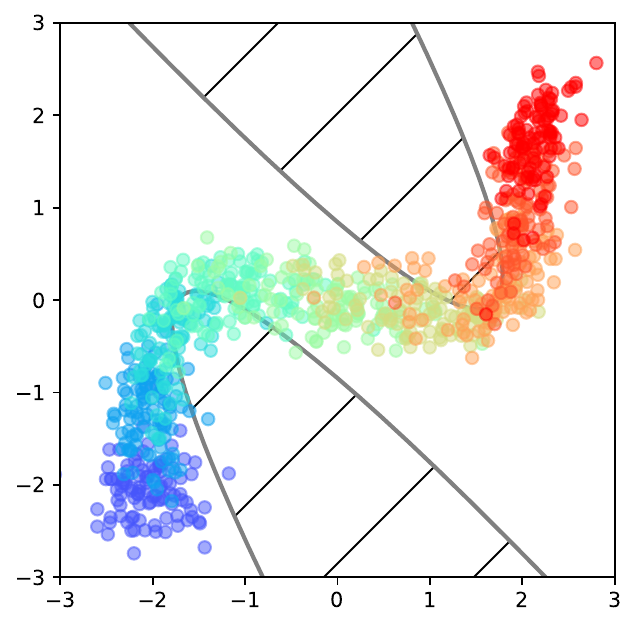}
		\end{tabular}
		\begin{tabular}{cc}
			\\
			\multicolumn{2}{c}{Log HJB Residuals}
			\\
			$\nu = 0$ & $\nu = 0.4$
			\\
			\includegraphics[width=0.3\textwidth]{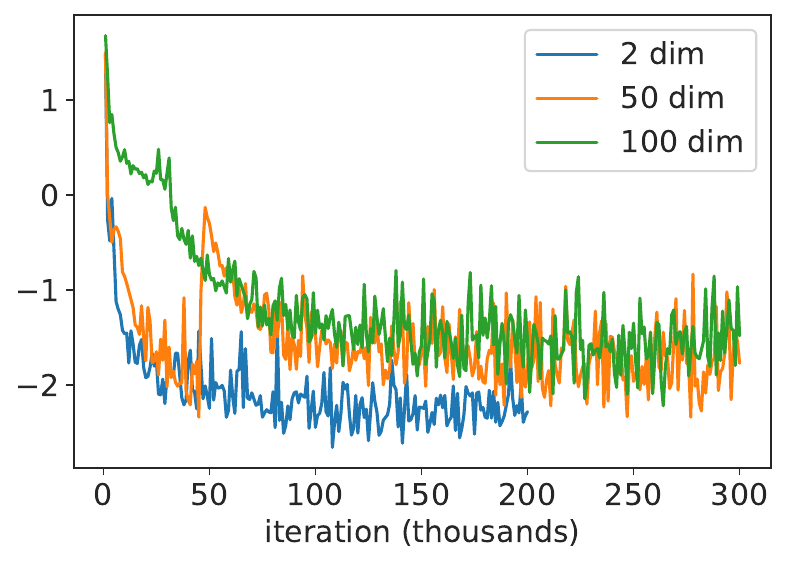}
			&
			\includegraphics[width=0.3\textwidth]{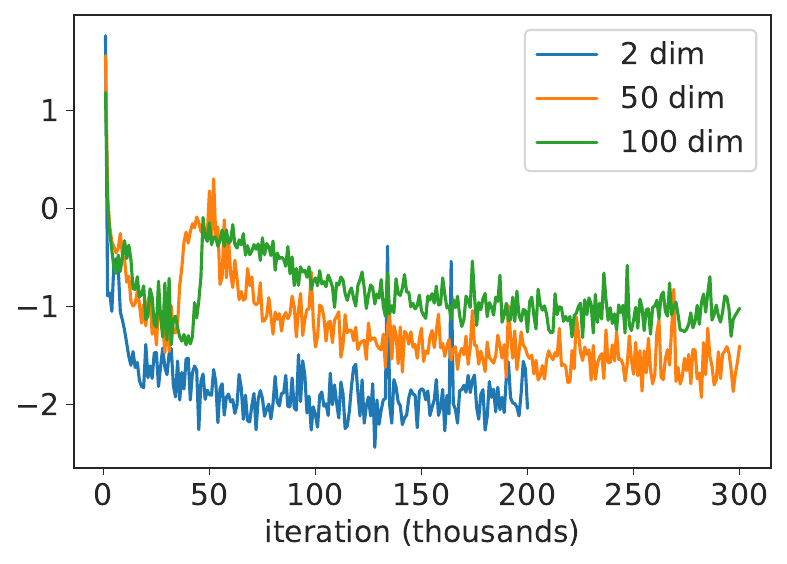}
		\end{tabular}
		\caption{Computation of the obstacle problem in dimensions $2$, $50$, and $100$ with stochasticity parameter $\nu = 0$ and $0.4$. For dimension $50$ and $100$, we plot the first two dimensions.}
		\label{fig:high-dim_obstacle}
	\end{figure}
	
	\begin{figure}[t]
		\centering
		\setlength{\tabcolsep}{0.5pt}
		\small
		\begin{tabular}{ccccc}
			\ & $d=2$ & $d=50$ & $d=100$  \ \\
			\rottext{$\nu = 0$} & \includegraphics[width=0.2\textwidth]{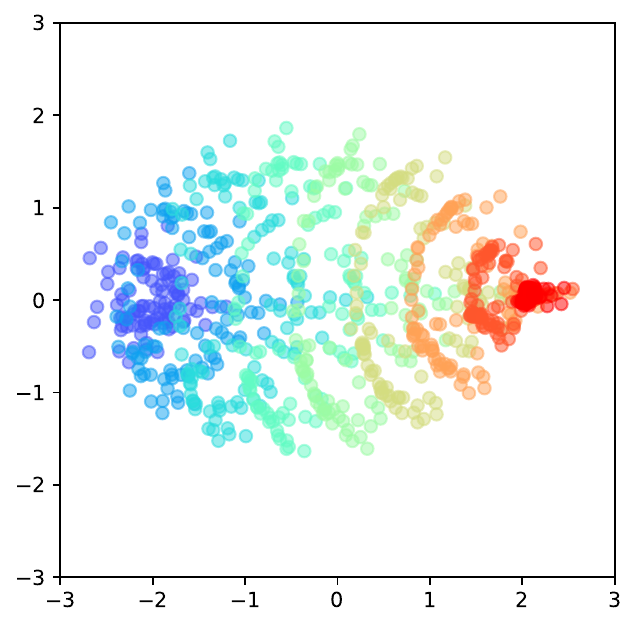}
			&
			\includegraphics[width=0.2\textwidth]{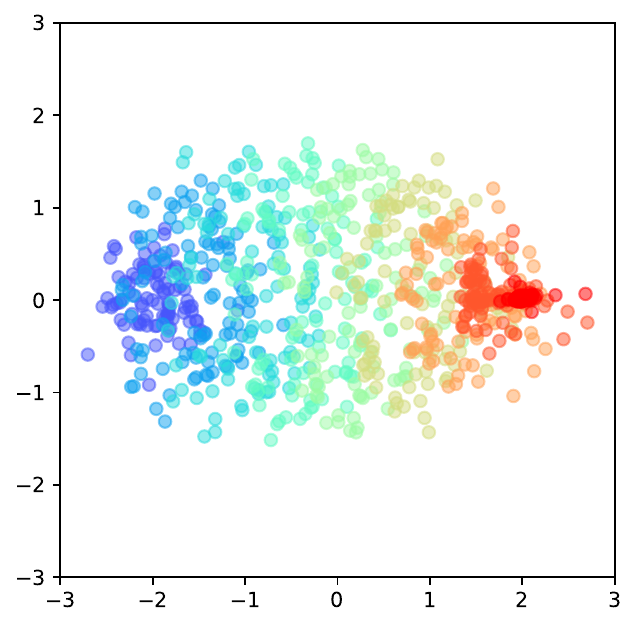}
			&
			\includegraphics[width=0.2\textwidth]{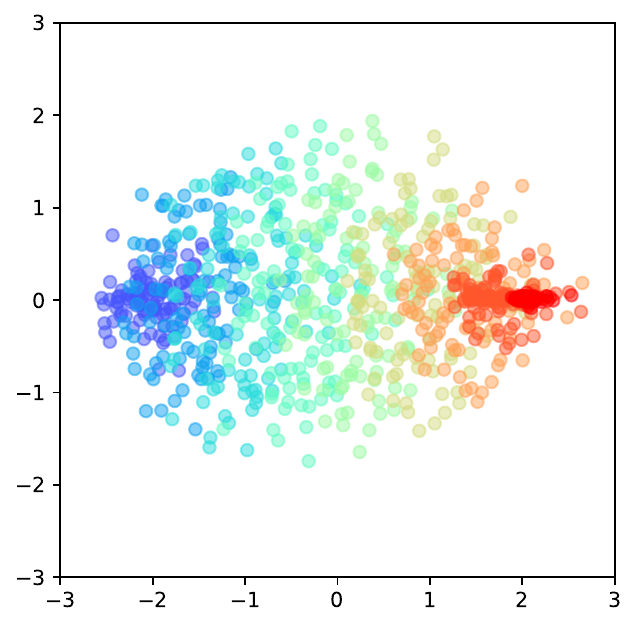} 
			\\
			\rottext{$\nu=0.4$} & \includegraphics[width=0.2\textwidth]{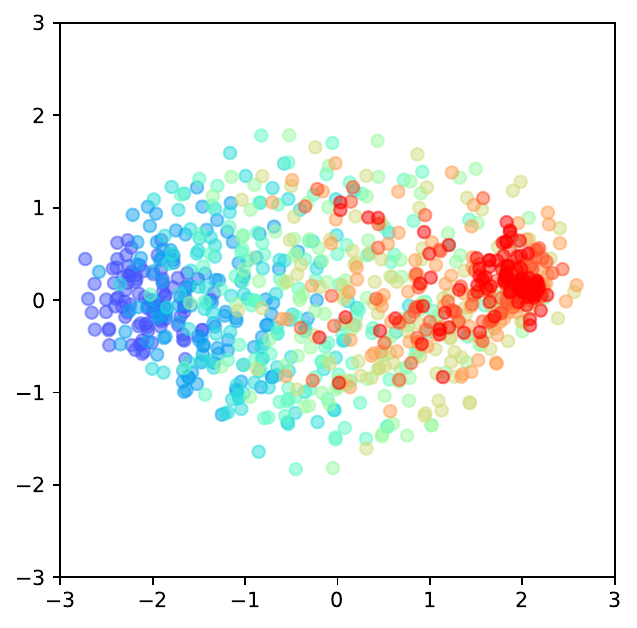}
			& 
			\includegraphics[width=0.2\textwidth]{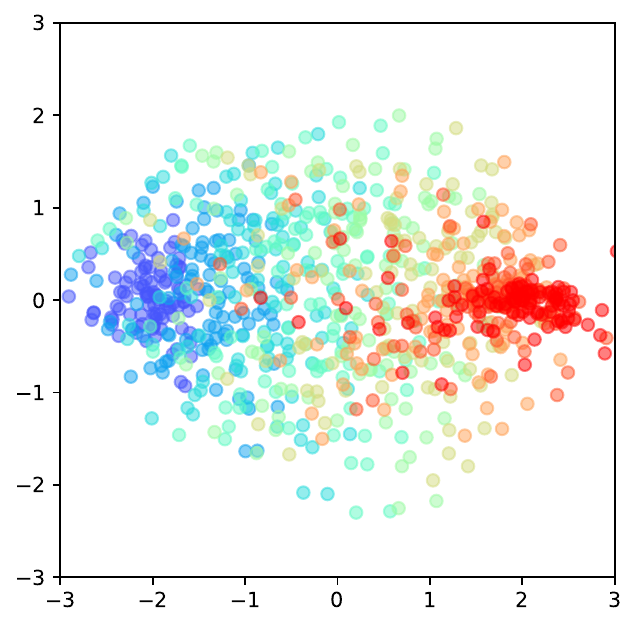}
			& 
			\includegraphics[width=0.2\textwidth]{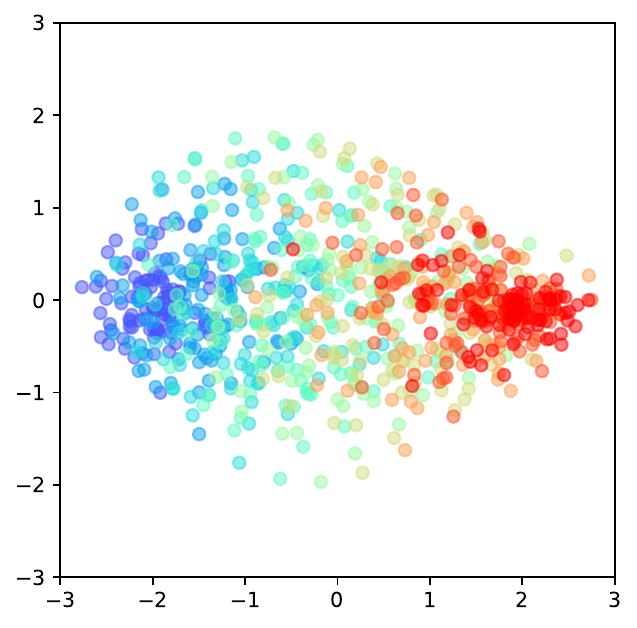}
		\end{tabular}
		\begin{tabular}{cc}
			\\
			\multicolumn{2}{c}{Log HJB Residuals}
			\\
			$\nu = 0$ & $\nu = 0.4$
			\\
			\includegraphics[width=0.3\textwidth]{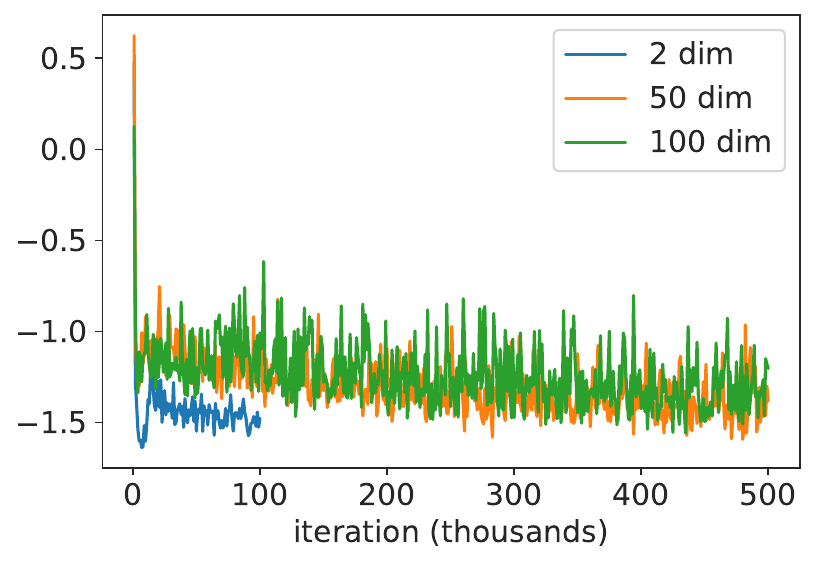}
			&
			\includegraphics[width=0.3\textwidth]{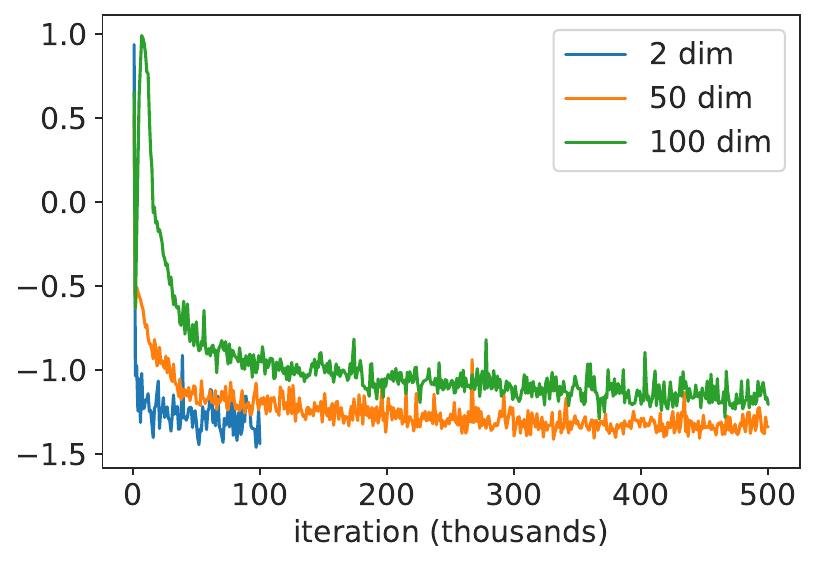}
		\end{tabular}
		\caption{Computation of the congestion problem in dimensions $2$, $50$, and $100$ with stochasticity parameter $\nu = 0$ and $0.4$. For dimensions $50$ and $100$, we plot the first two dimensions.}
		\label{fig:high-dim_congestion}
	\end{figure}
	\begin{figure}[t]
		\centering
		\setlength{\tabcolsep}{0.5pt}
		\small
		\begin{tabular}{ccccc}
			\ & $d=2$ & $d=50$ & $d=100$\\
			\rottext{$\nu = 0$} & \includegraphics[width=0.2\textwidth]{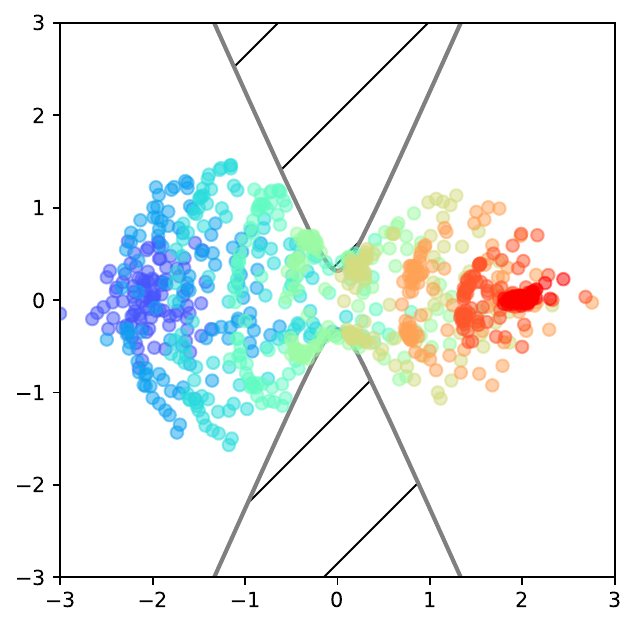}
			&
			\includegraphics[width=0.2\textwidth]{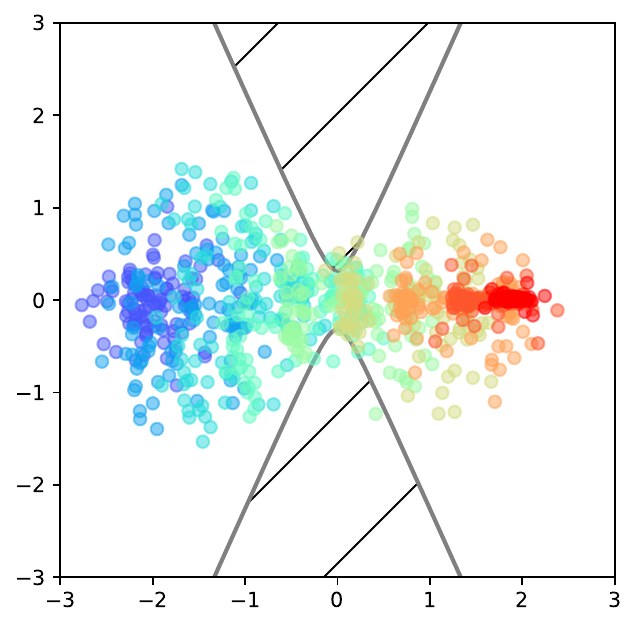}
			&
			\includegraphics[width=0.2\textwidth]{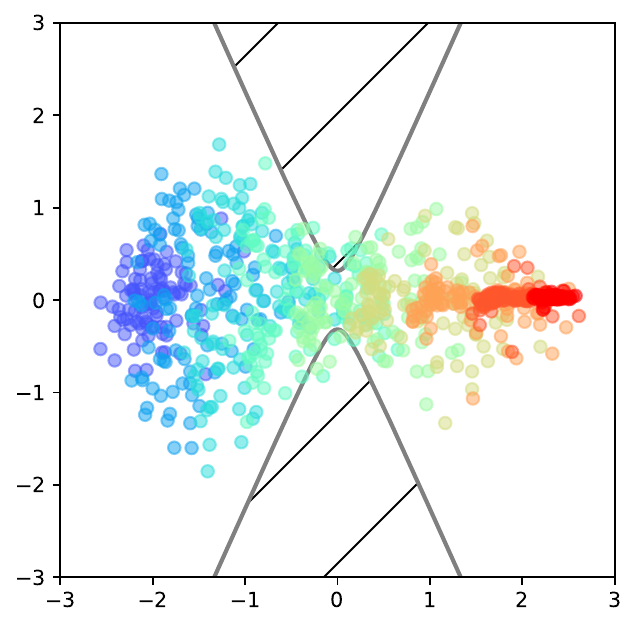}
			\\
			\rottext{$\nu = 0.1$} & \includegraphics[width=0.2\textwidth]{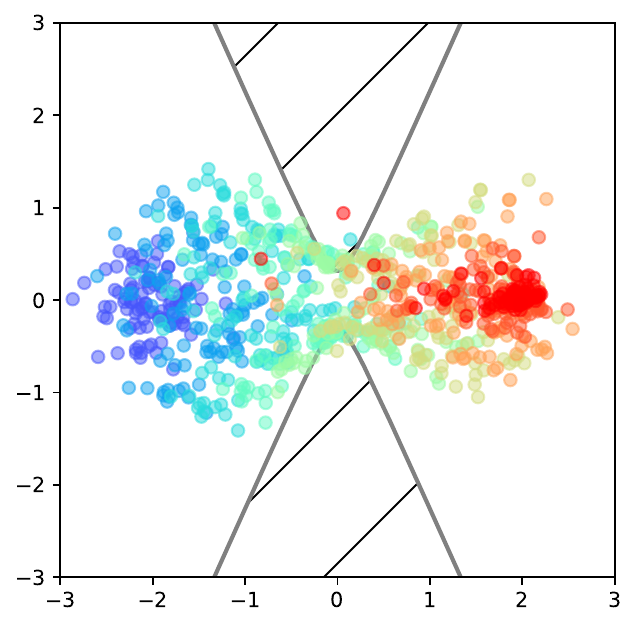}
			& 
			\includegraphics[width=0.2\textwidth]{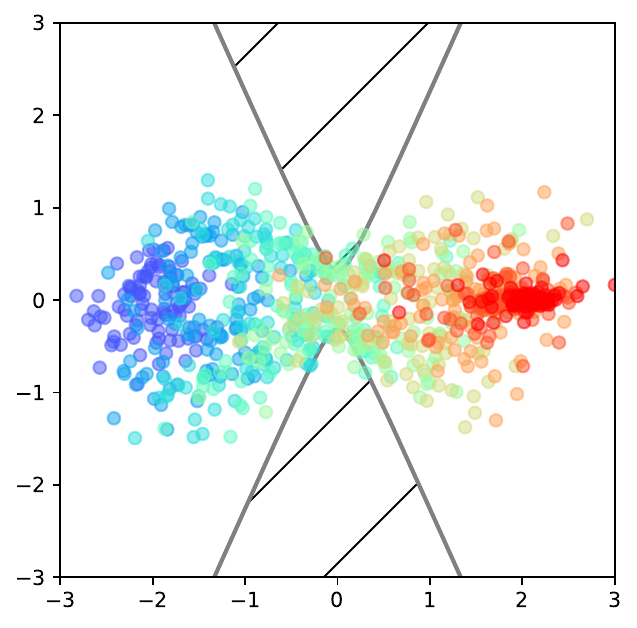}
			& 
			\includegraphics[width=0.2\textwidth]{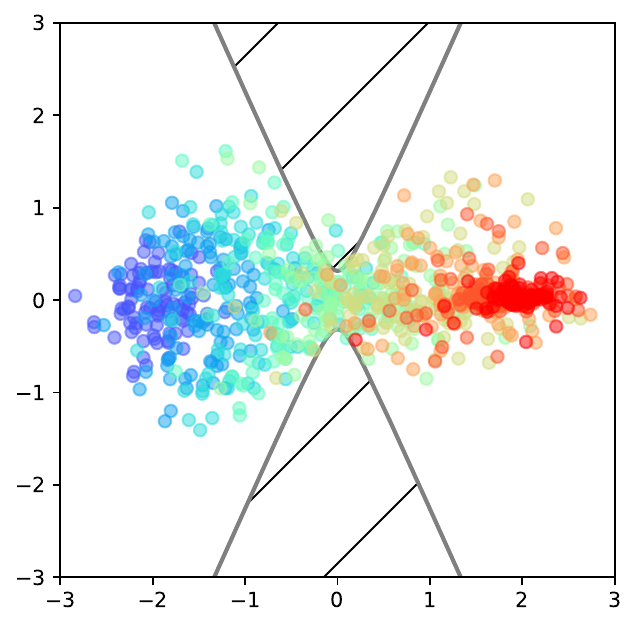}
		\end{tabular}
		\begin{tabular}{cc}
			\\
			\multicolumn{2}{c}{Log HJB Residuals}
			\\
			$\nu = 0$ & $\nu = 0.4$
			\\
			\includegraphics[width=0.3\textwidth]{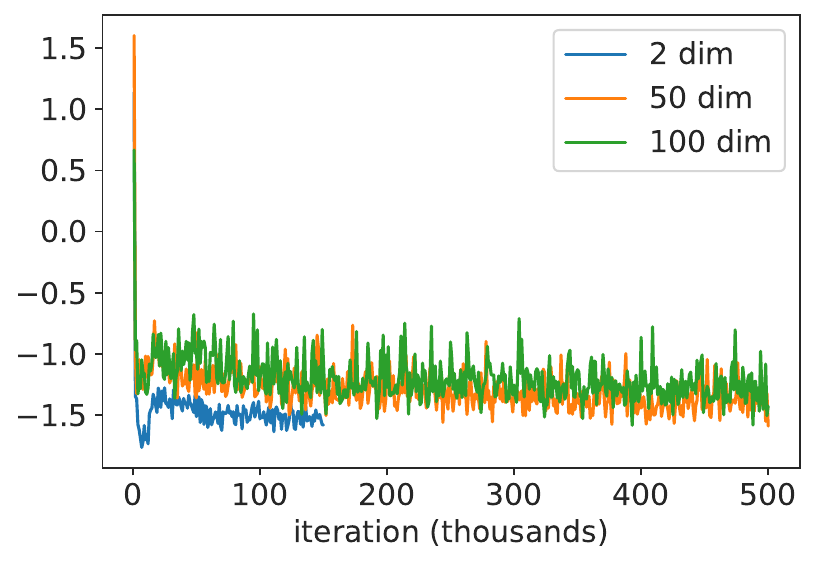}
			&
			\includegraphics[width=0.3\textwidth]{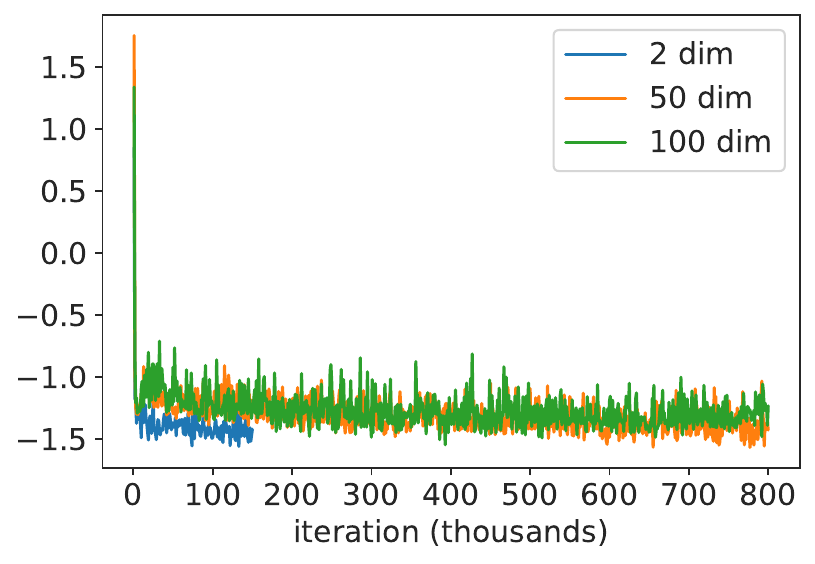}
		\end{tabular}
		\caption{Computation of the congestion problem with a bottleneck in dimensions $2$, $50$, and $100$ with stochasticity parameter $\nu = 0$ and $0.1$. For dimensions $50$ and $100$, we plot the first two dimensions.}
		\label{fig:high-dim_congestion_bottleneck}
	\end{figure}
	
	\begin{figure}[t]
		\centering
		\setlength{\tabcolsep}{2pt}
		\small
		\begin{tabular}{cc}
			Log relative error $(\gamma=0)$ & Log relative error $(\gamma=0.1)$
			\\
			\includegraphics[width=0.4\textwidth]{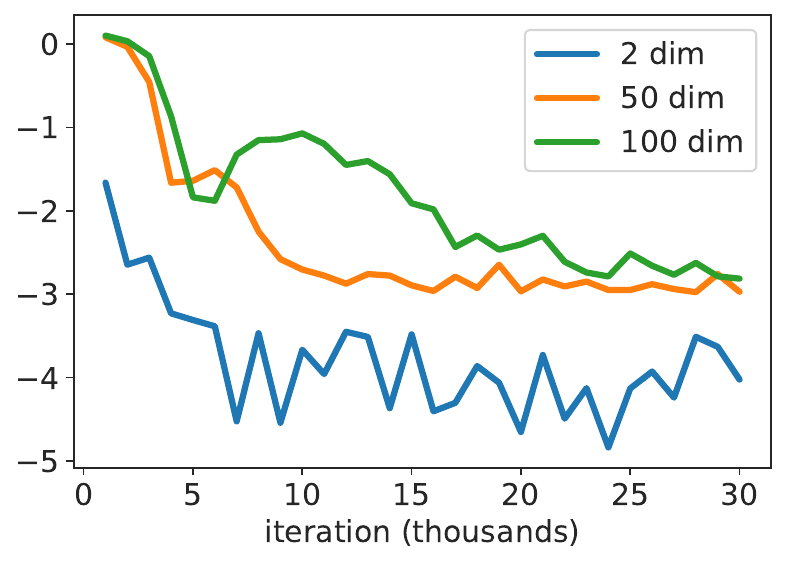} 
			&
			\includegraphics[width=0.4\textwidth]{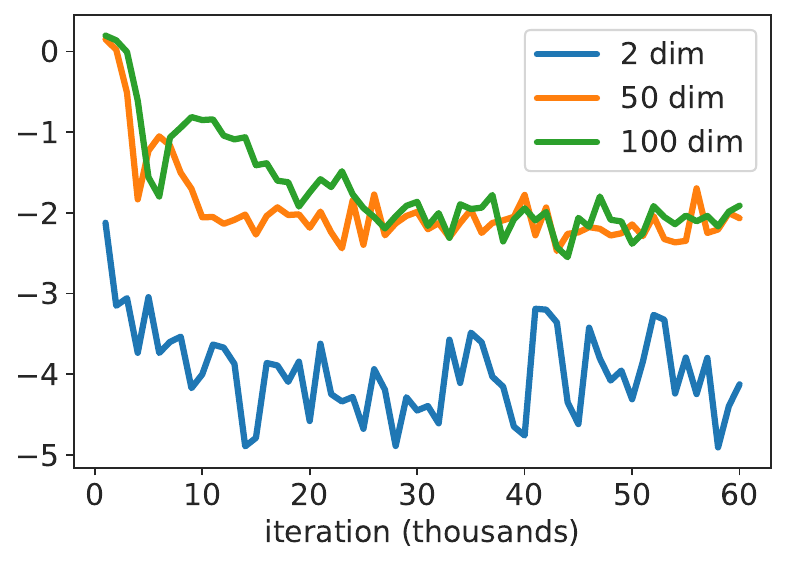} 
		\end{tabular}
		\caption{log relative errors in $2$, $50$, and $100$ dimensions, and for $\gamma=0, 0.1$. Here, $\gamma=0$ means no interaction. For the $d=2$ case, we compute the validation on a $32 \times 32$ grid over $16$ uniformly-spaced timesteps with the true $\phi$ from eq.~(\ref{eq:true_sol}). For the $d=50$ and $100$ case, we compute on a sample of 4096 sample points, sampled from the initial density.}
		\label{fig:analyticComparison}
	\end{figure}
	
	\begin{figure}[t]
		\centering
		\setlength{\tabcolsep}{0.5pt}
		\small
		\begin{tabular}{ccccc}
			\ & $\text{No Congestion}$ & $\text{With Congestion}$ \\
			\rottext{$\sigma = 0$} & \includegraphics[width=0.33\textwidth]{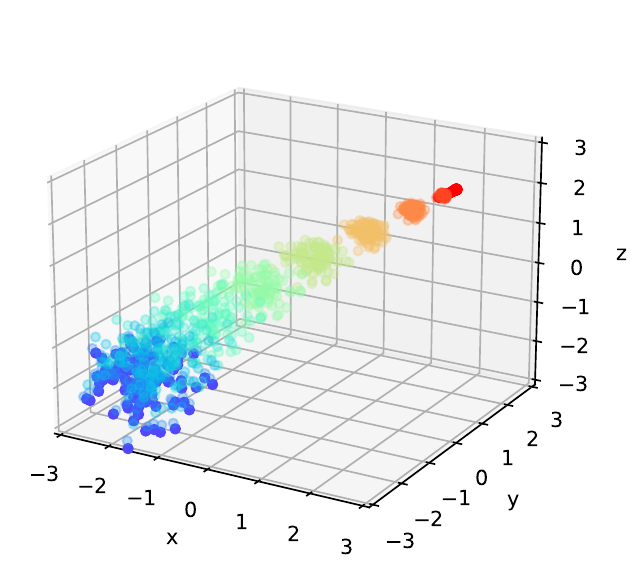}
			&
			\includegraphics[width=0.33\textwidth]{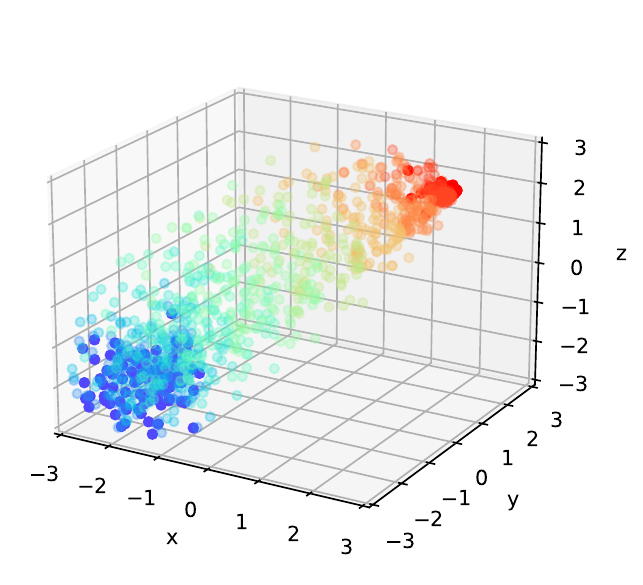}
			\\
			\rottext{$\sigma = 0.1$}  & \includegraphics[width=0.33\textwidth]{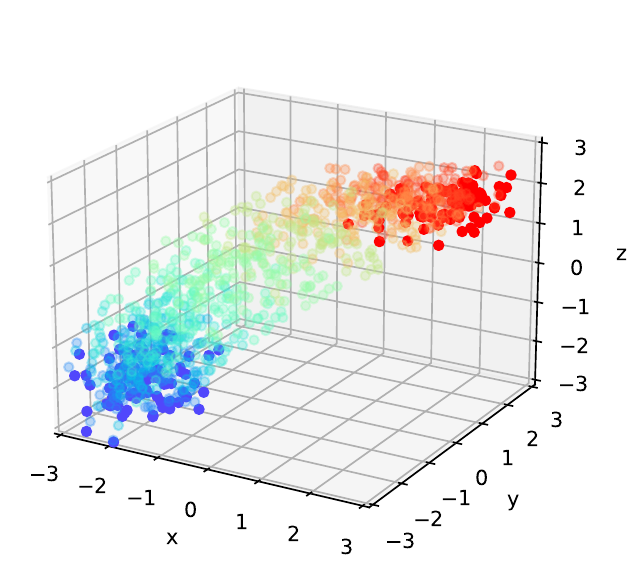}
			& \includegraphics[width=0.33\textwidth]{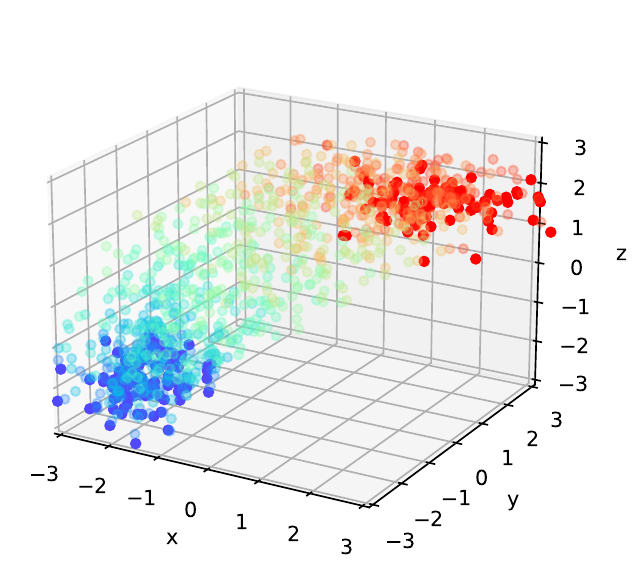}
		\end{tabular}
		\begin{tabular}{c}
			\\
			Log HJB Residuals
			\\
			\includegraphics[width=0.4\textwidth]{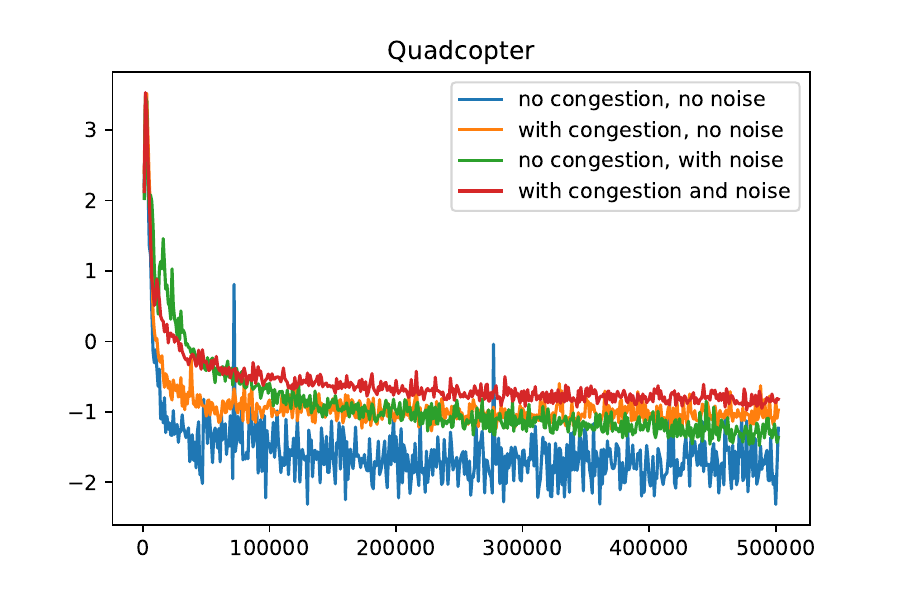}
		\end{tabular}
		\caption{Computation of the congestion problem with a bottleneck in dimensions $2$, $50$, and $100$ with stochasticity parameter $\nu = 0$ and $0.1$. For dimensions $50$ and $100$, we plot the first two dimensions.}
		\label{fig:quadcopter}
	\end{figure}
	
	\section{Numerical Experiments} \label{sec:experiments}
	We demonstrate the potential of APAC-Net on a series of high-dimensional MFG problems. We also illustrate the behavior the MFG solutions for different values of $\nu$ and use an analytical solution to illustrate the accuracy of APAC-Net.
	We also provide additional high-dimensional results in App.~\ref{app:B}.
	\paragraph{Experimental Setup}
	We assume without loss of generality $T=1$. In all experiments, our neural networks have three hidden layers, with $100$ hidden units per layer. We use a residual neural network (ResNet) for both networks, with skip connection weight $0.5$. For $\phi_\omega$, we use the Tanh activation function, and for $G_\theta$, we use the ReLU activation function. For training, we use ADAM with $\beta=(0.5, 0.9)$, learning rate $4\times 10^{-4}$ for $\phi_\omega$, learning rate $1\times 10^{-4}$ for $G_\theta$, weight decay of $10^{-4}$ for both networks, batch size $50$, and $\lambda=1$ (the HJB penalty parameter) in Algorithm~\ref{alg:APACnet}.
	
	The Hamiltonians in our experiments have the form
	\begin{equation}\label{eq:exp_ham}
	H(x,p, t) = c \|p\|_2 + f(x, \rho(x, t)),
	\end{equation}
	where $f(x, \rho(x, t))$ varies with the environment (either avoiding obstacles, or avoiding congestion, etc.), and $c$ is a constant (that represents maximal speed). Furthermore, we choose as terminal cost 
	\begin{equation}
	\sG(\rho(\cdot,T)) = \int_{\Omega}\|x - x_T\|_2 \rho(x,T)dx,
	\end{equation}
	which is the distance between the population and a target destination.
	To allow for verification of the high-dimensional solutions, we set the obstacle and congestion costs to only affect the first two dimensions. 
	In Figs.~\ref{fig:compareNu},~\ref{fig:high-dim_obstacle},~\ref{fig:high-dim_congestion}, and~\ref{fig:high-dim_congestion_bottleneck}, time is represented by color. Specifically, blue denotes starting time, red denotes final time, and the intermediate colors denote intermediate times.
	We also plot the HJB residual error, that is, $\ell_{\rm HJB}$ in Alg.~\ref{alg:APACnet}, on 4096 fixed sampled points which helps us monitor the convergence of APAC-Net.
	As in standard machine learning methods, all the plots in this section are generated using \emph{validation} data, i.e., data not used in training, in order to gauge generalizability of APAC-Net.
	Further details as well as additional experiments can be found in the appendix.
	
	\paragraph{Effect of Stochasticity Parameter $\nu$}
	\label{subsec:nuEffect}
	We investigate the effect of the stochasticity parameter $\nu$ on the behavior of the MFG solutions. In Fig.~\ref{fig:compareNu}, we show the solutions for 2-dimensional MFGs using $\nu = 0, 0.2, 0.4,$ and $0.6$.
	As $\nu$ increases, the density of agents widens along the paths due to the added the added diffusion term in the HJB and FP equations in~\eqref{eq:MFGPDEs}. 
	These results are consistent with those in~\cite{parkinson2020model}.
	
	\paragraph{Obstacles}
	We compute the solution to a MFG where the agents are required to avoid obstacles. In this case, we let
	\begin{equation}
	\begin{split}
	f(x_1, x_2, \ldots, x_d ) = & \; \gamma_{\text{obst}}(\max\{f_1(x_1, x_2), 0\} \\
	& + \max\{f_2(x_1, x_2),0\})
	\end{split}
	\end{equation}
	with $\gamma_\text{obst} = 5$, and denoting $R = \psm{\cos(\theta) & -\sin(\theta) \\ \sin(\theta) & \cos(\theta)}$ with $\theta=\pi/5$, $Q = \psm{5 & 0 \\ 0 & 0}$, and $b = (0, 2)$, then
	\begin{equation}
	\begin{split}
	f_1(x_1, x_2) &= -v^\top Qv - b\cdot v - 1, \\
	\text{with } v &= ((x_1, x_2) - (-2, 0.5))R.
	\end{split}
	\end{equation}
	Similarly, we let
	\begin{equation}
	\begin{split}
	f_2(x_1, x_2) &= -w^\top Qw + b\cdot w - 1, \\
	\text{with } w &= ((x_1, x_2) - (2, -0.5))R.
	\end{split}
	\end{equation}
	The obstacles $f_1$ and $f_2$ are shown in hatched markings in Fig.~\ref{fig:high-dim_obstacle}. 
	Our initial density $\rho_0$ is a Gaussian centered at $(-2, -2, 0, \ldots, 0)$ with standard deviation $1/\sqrt{10} \approx 0.32$, and the terminal function is $g(x) = \|(x_1, x_2) - (2,2)\|_2$. We chose $c=8$ in \eqref{eq:exp_ham}. The numerical results are shown in Fig.~\ref{fig:high-dim_obstacle}. Observe that results are similar across dimensions, verifying our high-dimensional computation.
	We run the two-dimensional problem for 200k iterations and the 50 and 100-dimensional problems for 300k iterations.
	
	\paragraph{Congestion}
	We choose the interaction term to penalize congestion, so that the agents are encouraged to spread out. In particular, we have
	\begin{equation}\label{eq:cong_pen}
	\mathcal{F}(\rho(x,t)) = \int_\Omega \int_\Omega \frac{1}{\|(x_1, x_2) - (y_1, y_2)\|^2 + 1} \,d\rho(x,t) \,d\rho(y,t),
	\end{equation}
	which is the (bounded) inverse squared distance, averaged over pairs of agents. Computationally, we sample from $\rho$ twice and then calculate the integrand. Here, our initial density $\rho_0$ is a Gaussian centered at $(-2, 0, -2, \ldots, -2)$ with standard deviation $1/\sqrt{10} \approx 0.32$, the terminal function is $\mathcal{G}(x) = \|(x_1, x_2) - (2,0)\|_2$, and we chose $c=5$ in \eqref{eq:exp_ham}. Results are shown in Fig.~\ref{fig:high-dim_congestion}, where we see qualitatively similar results across dimensions.
	We run the two-dimensional problem for 100k iterations and the 50 and 100-dimensional problems for 500k iterations.
	
	\paragraph{Congestion with Bottleneck Obstacle}
	We combine the congestion problem with a bottleneck obstacle. The congestion penalization is the same as~\eqref{eq:cong_pen}, and the obstacle represents a bottleneck -- thus agents are encouraged to spread out, but must squeeze together to avoid the obstacle. The initial density, terminal functions, $c$ in \eqref{eq:exp_ham}, and the expression penalizing congestion are the same as in the congestion experiment above. The obstacle is chosen to be
	\begin{equation}
	\begin{split}
	& f(v) = \gamma_{\text{obst}} \max\lt\{-v^\top \pmat{5 & 0 \\ 0 & -1} v - 0.1, 0\rt\} \text{ with } v = (x_1, x_2)
	\end{split}
	\end{equation}
	with $\gamma_\text{obst}=5$. As intuitively expected, the agents spread out before and after the bottleneck, but squeeze together in order to avoid the obstacle (see Fig.~\ref{fig:high-dim_congestion_bottleneck}). We run the two-dimensional problem for 100k iterations and the 50 and 100-dimensional problems for 500k iterations. We observe similar results across dimensions.
	
	\paragraph{Analytic Comparison}
	As a last experiment, we verify our method by comparing it to an analytic solution for dimensions 2, 50, and 100 with congestion $(\gamma = 0.1)$ and without congestion $(\gamma = 0)$. For 
	\begin{equation}
	\begin{split}
	f = \gamma \ln(\rho), \qquad H(x,p) = \frac{\| p \|^2}{2} - \frac{\beta \| x \|^2}{2}, \qquad g(x) = \frac{\alpha|x|^2}{2}-\left(\nu d \alpha +\frac{\gamma d}{2}\ln \frac{\alpha}{2\pi \nu}\right)
	\end{split}
	\end{equation}
	and $\nu = \beta = 1$ in~\eqref{eq:MFGPDEs}, the explicit formula for $\phi$ is given by
	\begin{equation}\label{eq:true_sol}
	\begin{split}
	\phi(x,t) = \frac{\alpha|x|^2}{2}-\left( d \alpha +\frac{\gamma d}{2}\ln \frac{\alpha}{2\pi}\right)t, \qquad \rho(x,t) = \left(\frac{\alpha}{2\pi}\right)^{\frac{d}{2}} e^{-\frac{\alpha|x|^2}{2}},
	\end{split}
	\end{equation}
	where $\alpha=\frac{-\gamma + \sqrt{\gamma^2 + 4}}{2}$. For the $\gamma=0.1$ case, we use Kernel Density Estimation \cite{rosenblatt1956, parzen1962} to estimate $\rho$ from samples of the generator. The derivation of the analytic solution can be found in App.~\ref{app:A}. 
	
	For $d=2$, we compute the relative error on a grid of size $32 \times 32 \times 16$, where we discretize the spatial domain $\Omega = [-2, 2]^2$ with $32 \times 32$ points and the time domain $[0 , 1]$ with $16$ points. 
	Note here that the grid points are \emph{validation} points, i.e., points that were not used in training.
	For $d=50$ and $d=100$, we use 4096 sampled points for validation since we cannot build a grid.
	We run a total of 30k and 60k iterations for $\gamma = 0$ and $\gamma = 0.1$, respectively. We validate every 1k iterations. Fig.~\ref{fig:analyticComparison} shows that our learned model approaches the true solution across all dimensions for both values of $\gamma$, indicating that APAC-Net generalizes well.

	\paragraph{A Realistic Example: The Quadcopter} 
	In this experiment, we examine a realistic scenario where the dynamics are that of a Quadcopter (a.k.a. Quadrotor craft), which is an aerial vehicle with four rotary wings (similar to many of the consumer drones seen today). The dynamics of the quadrotor craft are given as,
	\begin{equation*}
	\begin{split}
	\left\{ \begin{array}{rl}
	\ddot{x} &= \frac{u}{m}\l( \sin(\phi)\sin(\psi) + \cos(\phi)\cos(\psi)\sin(\theta) \r) \\
	\ddot{y} &= \frac{u}{m}\l(-\cos(\psi)\sin(\phi) + \cos(\phi)\sin(\theta)\sin(\psi) \r) \\
	\ddot{z} &= \frac{u}{m}\cos(\theta)\cos(\phi) - g \\
	\ddot{\psi} &= \tilde{\tau}_{\psi} \\
	\ddot{\theta} &= \tilde{\tau}_{\theta} \\
	\ddot{\phi} &= \tilde{\tau}_{\phi}
	\end{array} \right.
	\end{split}
	\end{equation*}
	where $x$, $y$, and $z$ are the usual Euclidean spatial coordinates, and $\phi$, $\theta$ and $\psi$ are the angular coordinates of roll, pitch, and yaw, respectively. The constant $m$ is the mass, to which we put $0.5$ (kg), and $g$ is the gravitation acceleration constant on Earth, to which we put $9.81$ (meters per second squared). The variables $u$, $\tilde{\tau}_{\psi}$, $\tilde{\tau}_{\theta}$, $\tilde{\tau}_{\phi}$ are the controls representing thrust, and angular acceleration, respectively. In order to fit a control framework, the above second-order system is turned into a first-order system: 
	\begin{equation*}
	\begin{split}
	\left\{ \begin{array}{rl}
	\dot{x}_1 &= x_2 \\
	\dot{x}_2 &= \frac{u}{m}\l( \sin(\phi_1)\sin(\psi_1) + \cos(\phi_1)\cos(\psi_1)\sin(\theta_1) \r) \\
	\dot{y}_1 &= y_2 \\
	\dot{y}_2 &= \frac{u}{m}\l(-\cos(\psi_1)\sin(\phi_1) + \cos(\phi_1)\sin(\theta_1)\sin(\psi_1) \r) \\
	\dot{z}_1 &= z_2 \\
	\dot{z}_2 &= \frac{u}{m}\cos(\theta_1)\cos(\phi_1) - g \\
	\dot{\psi}_1 &= \psi_2 \\
	\dot{\psi}_2 &= \tilde{\tau}_{\psi} \\
	\dot{\theta}_1 &= \theta_2 \\
	\dot{\theta}_2 &= \tilde{\tau}_{\theta} \\
	\dot{\phi}_1 &= \phi_2 \\
	\dot{\phi}_2 &= \tilde{\tau}_{\phi}
	\end{array} \right.
	\end{split}
	\end{equation*}
	which we will compactly denote as $\dot{\textbf{x}} = \textbf{h}(\textbf{x}, \textbf{u})$, where $\textbf{h}$ is the right-hand side, $\textbf{x}$ is the state, and $\textbf{u}$ is the control. As can be observed, the above is a 12-dimensional system that is highly nonlinear and high-dimensional. In the stochastic case, we add a noise term to the dynamics: $d\textbf{x} = \textbf{h}(\textbf{x}, \textbf{u})\,dt + \sigma\, d\textbf{W}_t$, where $\textbf{W}$ denotes a Wiener process (standard Brownian motion). The interpretation here is that we are modeling the situation when the quadcopter suffers from noisy measurements.
	
	In our experiments, we set our Lagrangian cost function to be $L(\textbf{u}) = \frac{1}{2}\|\textbf{u}\|_2^2$ and thus our Hamiltonian becomes $H(\textbf{p}) = \frac{1}{2}\|\textbf{p}\|_2^2$. Our initial density $\rho_0$ is a Gaussian in the spatial coordinates $(x,y,z)$ centered at $(-2,-2,-2)$ with standard deviation $0.5$, and we set all other initial coordinates to zero (i.e. initial velocity, initial angular position, and initial angular velocity are all set to zero). We set our terminal cost to be a simple norm difference between the agents current position and the position $(2,2,2)$, and we also want the agents to have zero velocity, i.e.
	\begin{equation*}
	\sG(\rho(\cdot,T)) = \int_{\Omega}\|(x, y, z, \dot{x}, \dot{y}, \dot{z}) - (2, 2, 2, 0, 0, 0)\|_2 \rho(x,T)dx.
	\end{equation*}
	In our experiments we set the final time to be $T=4$.
	
	We also add a congestion term to our experiments where the congestion is in the spatial positions, so as to encourage agents to spread out (and thus making mid-air collisions less likely):
	\begin{equation*}
	\begin{split}
	&\mathcal{F}(\rho(x,y,z)) = \gamma_{\text{cong}}\frac{1}{(2\pi)^{\frac{3}{2}}} \int_\Omega \int_\Omega e^{\left(-\frac{1}{2} \|(x,y,z) - (\hat{x},\hat{y},\hat{z})\|_2^2\right)}\,d\rho((x,y,z))\,d\rho(\hat{x},\hat{y},\hat{z})
	\end{split}
	\end{equation*}
	where in our experiments we put $\gamma_{\text{cong}}=20$.
	
	For hyperparameters, we use the same setup as before, except we raise the batch size to 150. Our results are shown in Fig. \ref{fig:quadcopter}. We see that with congestion, the agents spread out more as expected. Furthermore, in the presence of noise, the agents' sensors are noisy, and so at terminal time the agents do no get as close to the terminal point of $(2,2,2)$. This noise also adds an envelope of uncertainty, so we do see the agents are not as streamlined as in the noiseless cases.

	\section{Conclusion}\label{sec:conclusion}
	We present APAC-Net, an alternating population-agent control neural network for solving high-dimensional stochastic mean field games. To this end, our algorithm avoids the use of spatial grids by parameterizing the controls, $\phi$ and $\rho$, using two neural networks, respectively. 
	Consequently, our method is geared toward high-dimensional instances of these problems that are beyond reach with existing grid-based methods. 
	APAC-Net therefore sets the stage for solving realistic high-dimensional MFGs arising in, e.g., economics~\cite{moll14,moll17,gueant2011mean,gomes2015economic}, swarm robotics~\cite{liu2018mean,elamvazhuthi2019mean}, and perhaps most important/relevant, epidemic modelling~\cite{lee2020controlling,chang2020game}.
	Our method also has natural connections with Wasserstein GANs, where $\rho$ acts as a generative network and $\phi$ acts as a discriminative network. Unlike GANs, however, APAC-Net incorporates the structure of MFGs via~\eqref{eq:minmax_MFG_gen} and~\eqref{eq:NNStructure}, which absolves the network from learning an entire MFG solution from the ground up. 
	Our experiments show that our method is able to solve 100-dimensional MFGs. 
	
	\edits{
		Since our method was presented solely in the setting of potential MFGs, a natural extension is the non-potential MFG setting where the MFG can no longer be written in a variational form. Instead, one would have to formulate the MFG as a monotone inclusion problem~\cite{liu2020splitting}. 
		Moreover, convergence properties of the training process of APAC-Net may be investigated following the techniques presented in, e.g.,~\cite{cao2020approximation}.}
	Finally, a practical direction involves examining guidelines on the design of more effective network architectures, e.g., PDE-based networks~\cite{haber2017stable,ruthotto2019deep}, neural ODEs~\cite{chen2018neural}, or sorting  networks~\cite{anil2019sorting}. 
	To promote access and progress, we provide our PyTorch implementation at \url{https://github.com/atlin23/apac-net}.
	
	\begin{ack}
		The authors are supported by AFOSR MURI FA9550-18-1-0502, AFOSR Grant No. FA9550-18-1-0167, and ONR Grant No. N00014-18-1-2527.
	\end{ack}
	
\clearpage
\newpage
	\appendix
	\section{Derivation of Analytic Solution}
	\label{app:A}
	
	We derive explicit formulas used to test our approximate solutions in Sec.~\ref{sec:experiments}. Assume that $\nu,\beta>0,\gamma\geq 0$ and
	\begin{equation}
	H(x,p,t)=\frac{|p|^2}{2}- \frac{\beta |x|^2}{2},\quad f(x,\rho)=\gamma \ln \rho.
	\end{equation}{}
	Then \eqref{eq:MFGPDEs} becomes
	\begin{equation}\label{eq:MFG_ln}
	\begin{split}
	-\partial_t \phi - \nu \Delta \phi + \frac{|\nabla \phi|^2}{2} - \frac{\beta |x|^2}{2}= \gamma \ln \rho,
	\\
	\partial_t \rho - \nu \Delta \rho - \text{div} (\rho \nabla \phi) = 0,
	\\
	\rho(x,0) = \rho_0, \quad \phi(x,T) = \Psi(x).
	\end{split}    
	\end{equation}
	We find solutions to this system by searching for stationary solutions first:
	\begin{equation}\label{eq:MFG_stat}
	\begin{split}
	- \nu \Delta \phi + \frac{|\nabla \phi|^2}{2} - \frac{\beta |x|^2}{2}= \gamma \ln \rho+\Bar{H},
	\\
	- \nu \Delta \rho - \text{div} (\rho \nabla \phi) = 0,
	\end{split}
	\end{equation}
	and then writing
	\begin{equation}
	\phi(x,t)=\phi(x)-t \Bar{H},\quad \rho(x,t)=\rho(x).
	\end{equation}
	The second equation in \eqref{eq:MFG_stat} yields $\rho=c e^{-\frac{\phi}{\nu}}$, where $c$ is chosen so that $\int \rho=1$. Plugging this in the first equation in \eqref{eq:MFG_stat} we obtain
	\begin{equation}
	- \nu \Delta \phi + \frac{|\nabla \phi|^2}{2} - \frac{\beta |x|^2}{2}=\gamma \ln c-\frac{\gamma}{\nu} \phi +\Bar{H}.
	\end{equation}
	Now we make an ansatz that $\phi(x)=\frac{\alpha |x|^2}{2}$. Then we have that $\Delta \phi=d \alpha ,~\nabla \phi = \alpha x$, and obtain
	\begin{equation}
	- \nu d \alpha  + \frac{\alpha^2|x|^2}{2} - \frac{\beta |x|^2}{2}=\gamma \ln c-\frac{\gamma \alpha |x|^2}{2\nu} +\Bar{H}.
	\end{equation}
	Therefore, we have that
	\begin{equation}
	\alpha^2+\frac{\gamma \alpha}{\nu}=\beta,\quad \Bar{H}=-\nu d \alpha -\gamma \ln c.
	\end{equation}
	From the first equation, we obtain that
	\begin{equation}\label{eq:alpha}
	\alpha=\frac{-\gamma+\sqrt{\gamma^2+4\nu^2 \beta}}{2\nu}.
	\end{equation}
	On the other hand, we have that
	\begin{equation}
	\int \rho = c \int e^{-\frac{\alpha |x|^2}{2\nu}}dx=c \left(\frac{2\pi \nu}{\alpha}\right)^{\frac{d}{2}}=1,
	\end{equation}
	so
	\begin{equation}
	c=\left(\frac{\alpha}{2\pi \nu}\right)^{\frac{d}{2}} 
	\quad \text{ and } \quad \Bar{H}=-\nu d \alpha -\frac{\gamma d}{2}\ln \frac{\alpha}{2\pi \nu}.
	\end{equation}
	Summarizing, we get that for any $\nu,\beta>0,\gamma\geq 0$ the following is a solution for \eqref{eq:MFG_ln}
	\begin{equation}\label{eq:MFG_explicit}
	\begin{split}
	\phi(x,t) &= \frac{\alpha|x|^2}{2}-\left(\nu d \alpha +\frac{\gamma d}{2}\ln \frac{\alpha}{2\pi \nu}\right)t, \\
	\rho(x,t) &= \left(\frac{\alpha}{2\pi \nu}\right)^{\frac{d}{2}} e^{-\frac{\alpha|x|^2}{2\nu}}
	\end{split}
	\end{equation}{}
	where $\alpha$ is given by \eqref{eq:alpha}, and
	\begin{equation}\label{eq:MFG_explicit_initial_terminal}
	\begin{split}
	g(x) &= \frac{\alpha|x|^2}{2}-\left(\nu d \alpha +\frac{\gamma d}{2}\ln \frac{\alpha}{2\pi \nu}\right)T, \\ 
	\rho_0(x) &= \left(\frac{\alpha}{2\pi \nu}\right)^{\frac{d}{2}} e^{-\frac{\alpha|x|^2}{2\nu}}.
	\end{split}
	\end{equation}
	Choosing $\beta = \nu = 1$, \eqref{eq:MFG_explicit} gives the analytic solution used in Sec.~\ref{sec:experiments}.
	
	\section{Details on Numerical Results and More Experiments}
	\label{app:B}
	
	\paragraph{Congestion}
	
	Here we elaborate on how we compute the congestion term,
	\begin{equation}
	\mathcal{F}(\rho(x,t)) = \int_\Omega \int_\Omega \frac{1}{\|(x_1, x_2) - (y_1, y_2)\|^2 + 1} \,d\rho(x,t) \,d\rho(y,t),
	\end{equation}
	We do this by first using the batch $\{z_b\}_{b=1}^B$, which was used for training (and sampled from $\rho_0$), and then compute another batch $\{y_b\}_{b=1}^B$ again sampled from $\rho_0$. Letting $\{t_b\}_{b=1}^B$ be a batch of time-points uniformly sampled in $[0,1]$, we estimate the interaction cost with,
	\begin{equation}
	\mathcal{F}(\rho(x,t)) \approx \sum_{i=1}^B \frac{1}{\|G_\theta(z_b, t_b) - G_\theta(y_b, t_b)\|^2 + 1}.
	\end{equation}
	
	\paragraph{Quadcopter}
	
	\paragraph{Congestion with Bottleneck Obstacle and Higher Stochasticity}
	
\begin{figure}[t]
		\centering
		\setlength{\tabcolsep}{0.5pt}
		\small
		\begin{tabular}{ccccc}
			\ & $d=2$ & $d=50$ & $d=100$ \\
			\rottext{$\nu = 0.4$} & \includegraphics[width=0.25\textwidth]{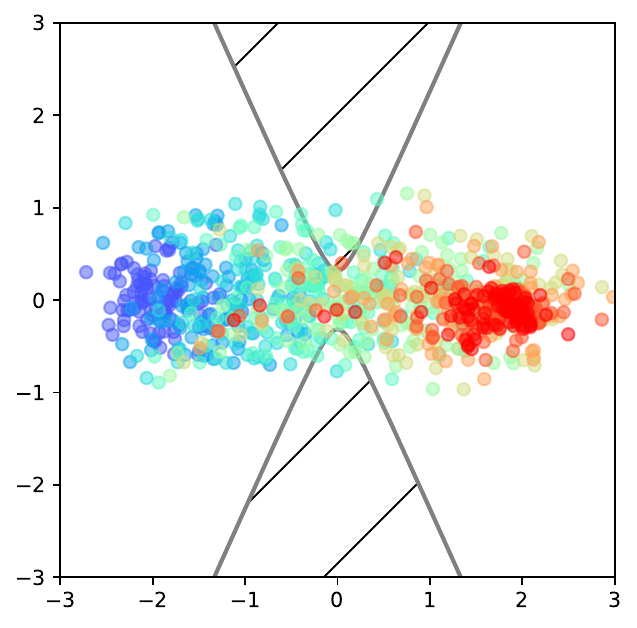}
			& 
			\includegraphics[width=0.25\textwidth]{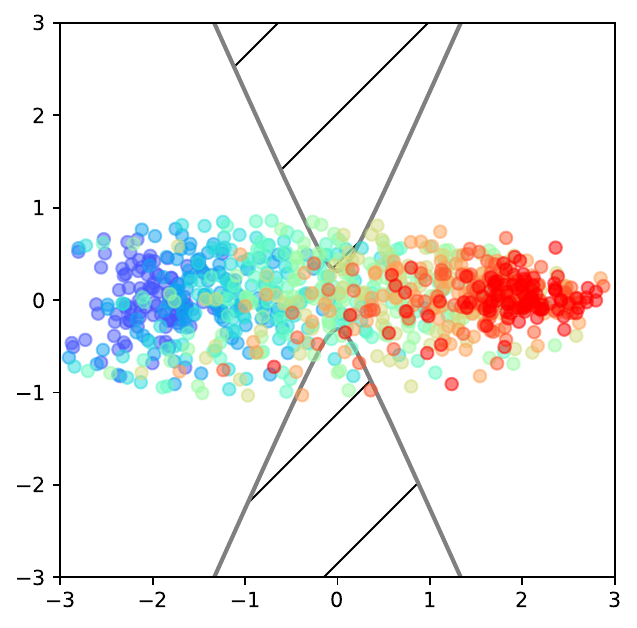}
			& 
			\includegraphics[width=0.25\textwidth]{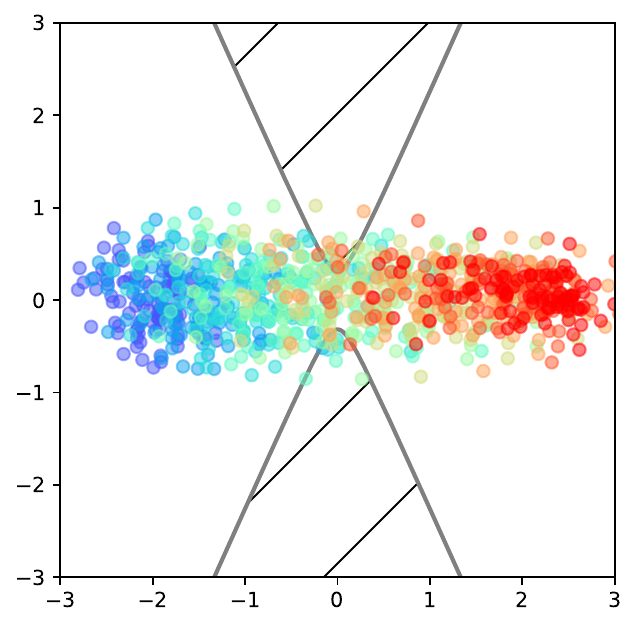}
		\end{tabular}
		\begin{tabular}{c}
			\\
			Log HJB Residuals $\nu = 0.4$
			\\
			\includegraphics[width=0.36\textwidth]{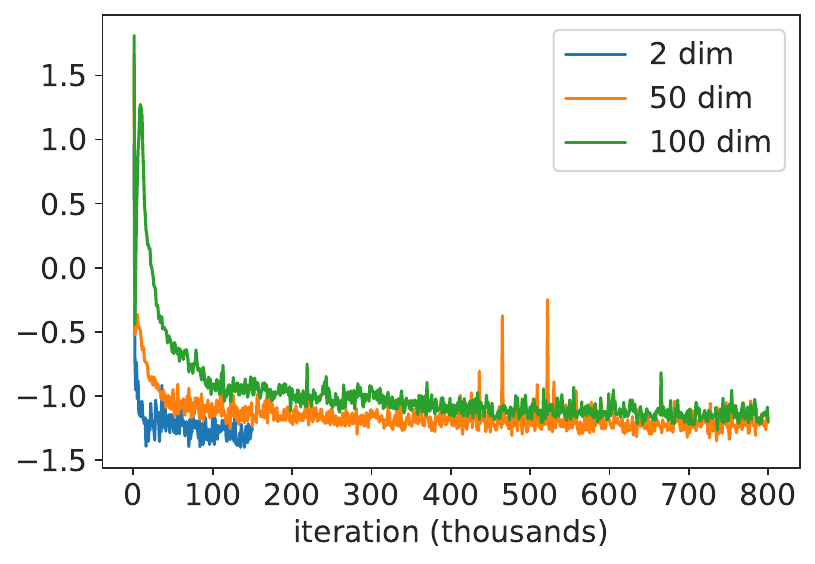}
		\end{tabular}
		\caption{Computation of the congestion problem with a bottleneck in dimensions $2$, $50$, and $100$ with stochasticity parameter $\nu = 0.4$. For dimensions $50$ and $100$, we plot the first two dimensions.}
		\label{fig:high-dim_congestion_bottleneck_nu-0p4}
	\end{figure}

	As mentioned in the main text, when choosing a stochasticity parameter $\nu > 0.1$, the stochasticity dominates the dynamics and the obstacles do not interact as much with the obstacle. 
	We plot these results in Fig.~\ref{fig:high-dim_congestion_bottleneck_nu-0p4}, where for $2$ dimensions, we trained for 150k iterations, and for $50$ and $100$ dimensions, we trained for 800k iterations.
	All environment and training parameters are the same as in the main text, except that now $\nu=0.4$.
	
	\paragraph{Analytic Comparison}
	
	Here we mention specifically how we performed Kernel Density Estimation. Namely, in order to estimate the density $\rho$, we take a batch of samples $\{z_b\}_{b=1}^B$ (during training, this is the training batch). Then at uniformly spaced time-points $\{t_b\}_{b=1} \subseteq [0,1]$, we estimate the density with the formula,
	\begin{equation}
	\rho(z_b, t_b) \approx \frac{1}{B} \frac{1}{(\sigma h \sqrt{2\pi})^d}  \sum_{i=1}^B \sum_{j=1}^B \text{exp}\lt( \frac{\|z_i - z_j\|^2 }{ (h\sigma)^2 } \rt)
	\end{equation}
	where we choose $\sigma=\sqrt{\frac{\gamma}{\nu}}$, and $d$ is the dimension, and $h=B^{-\frac{1}{d+4}}$, in accordance with Scott's rule for multivariate kernel density estimation~\cite{scott2005multidimensional}.
	
	\paragraph{Density Splitting via Symmetric Obstacle}
	Here we compute an example where we have a symmetric obstacle, and thus the generator will learn to split the density. Agents will go left or right of the obstacle depending on their starting position. Here we chose the obstacle as,
	\begin{equation}
	\begin{split}
	& f(x_1, x_2, \ldots, x_d) = \alpha_{\text{obst}} \max\lt\{ -v^\top Q v + 0.1, 0\rt\}, \\
	& Q=\pmat{1 & 0.8 \\ 0.8 & 1}, \;v=(x_1, x_2).
	\end{split}
	\end{equation}
	and we choose $\gamma_\text{obst} = 20$. The environment and training parameters are the same as in the main text, except we choose the HJB penalty $\lambda$ in Alg.~\ref{alg:APACnet} to be $0.1$. Qualitatively, we see the solution agrees with our intuition: the agents will go left or right depending on their starting position. Note that the results are similar across dimensions, verifying our computation. For the $2d$, $\nu=0$ case we trained for 100k iterations, for the $2d$, $\nu=0.1$ case we trained fro 300k iterations, for the $50d$ and $100d$, $\nu=0$ case we trained for 500k iterations, for the $50d$ $\nu=0.1$ case we trained for 1000k iterations, and the for the $100d$, $\nu=0.1$ case we trained for 2000k iterations.
	
	\begin{figure}[t]
		\centering
		\setlength{\tabcolsep}{0.5pt}
		\small
		\begin{tabular}{ccccc}
			\ & $d=2$ & $d=50$ & $d=100$ \\
			\rottext{$\nu = 0$} & \includegraphics[width=0.2\textwidth]{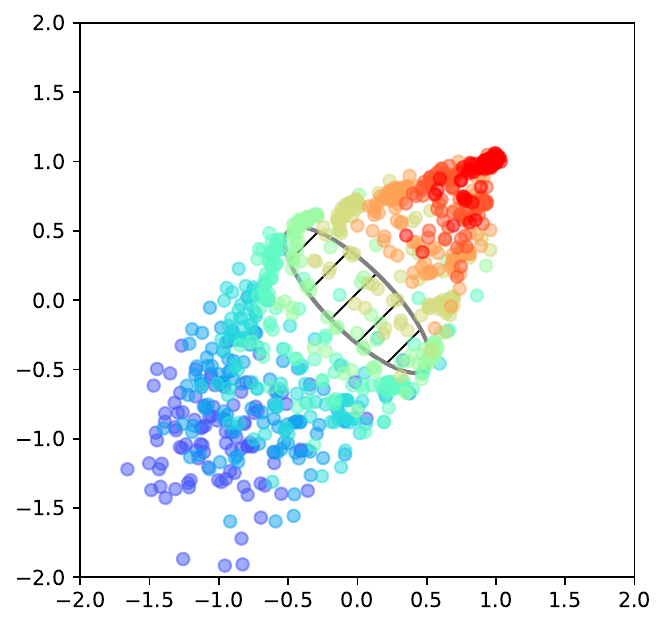}
			&
			\includegraphics[width=0.2\textwidth]{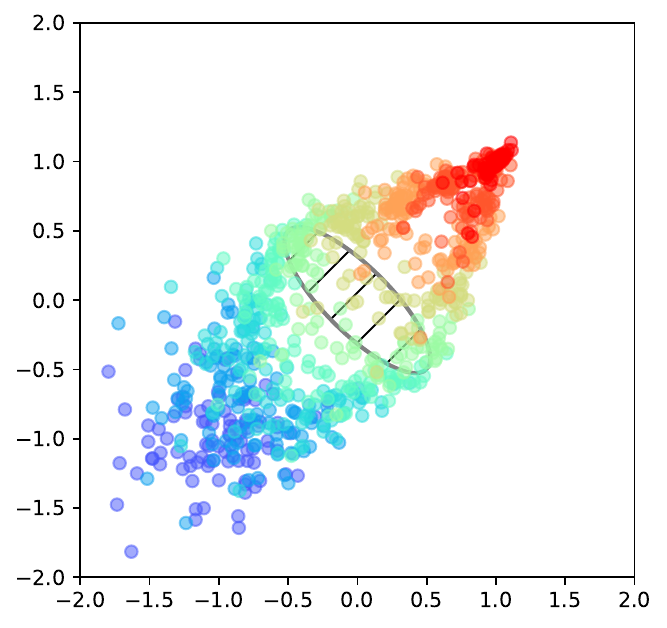}
			&
			\includegraphics[width=0.2\textwidth]{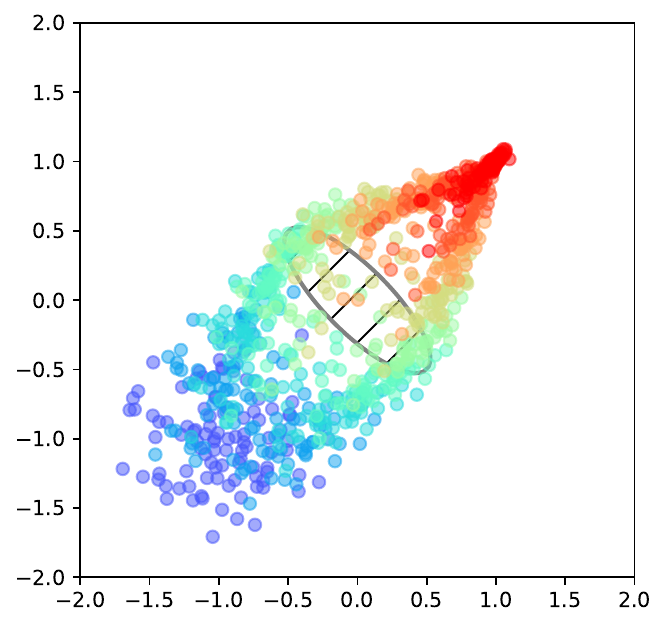}
			\\
			\rottext{$\nu = 0.1$} & \includegraphics[width=0.2\textwidth]{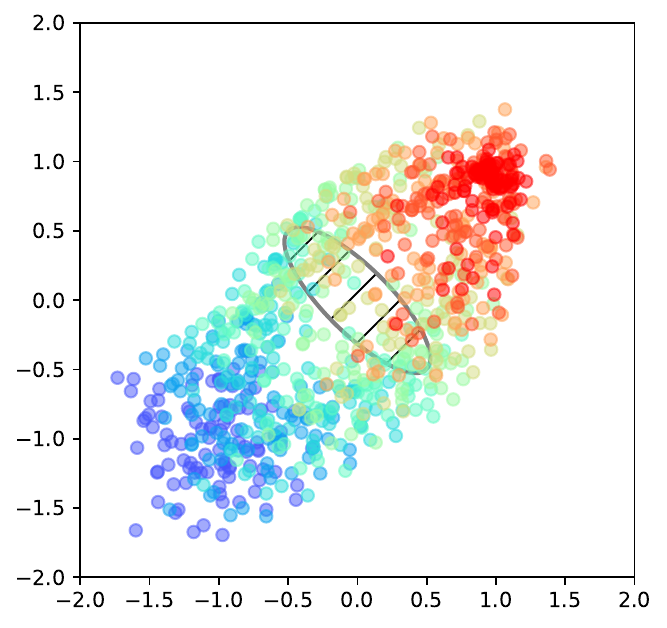}
			& 
			\includegraphics[width=0.2\textwidth]{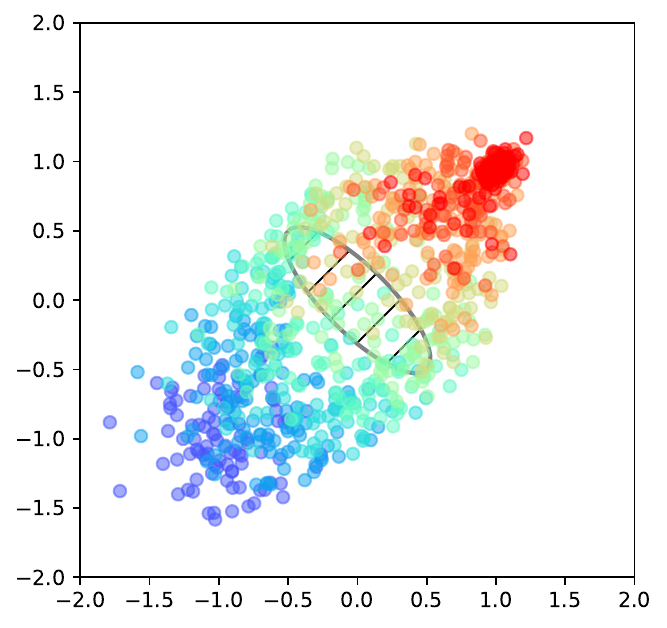}
			& 
			\includegraphics[width=0.2\textwidth]{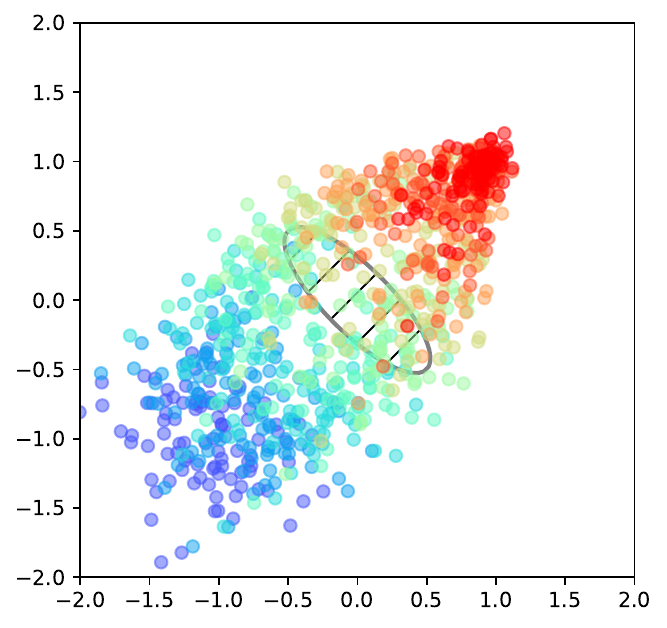}
		\end{tabular}
		\begin{tabular}{cc}
			\\
			\multicolumn{2}{c}{Log HJB Residuals}
			\\
			$\nu = 0$ & $\nu = 0.4$
			\\
			\includegraphics[width=0.3\textwidth]{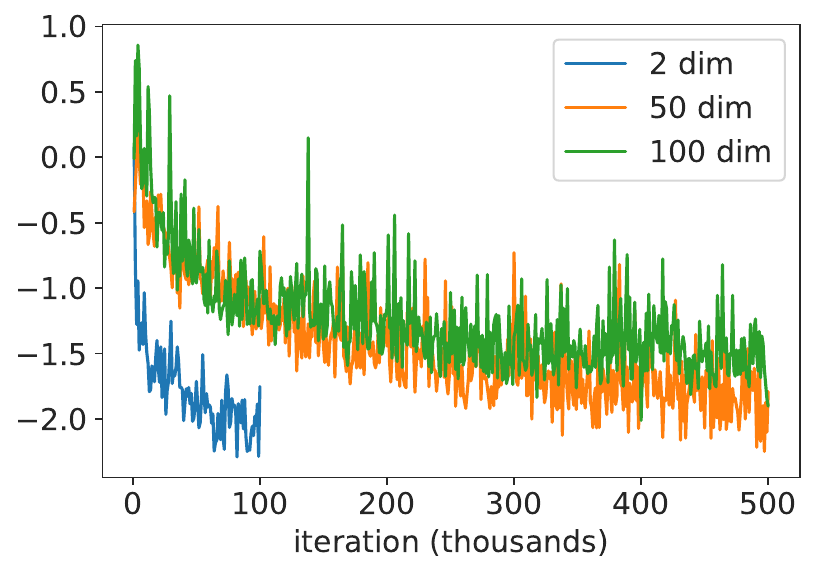}
			&
			\includegraphics[width=0.3\textwidth]{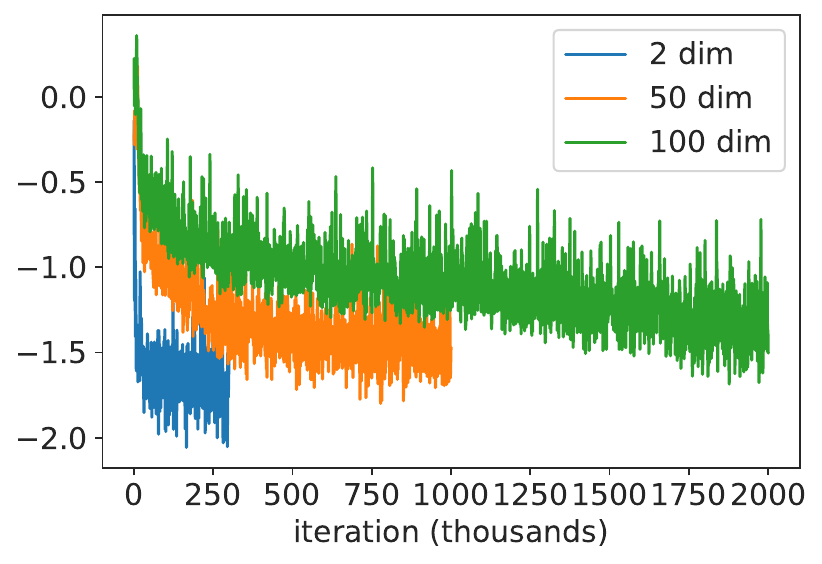}
		\end{tabular}
		\caption{Computation of the an obstacle problem where the obstacle is symmetric. We plot the results for dimensions $2$ and $100$, and for $\nu=0$ and $\nu=0.1$.}
		\label{fig:symmetric_obstacle}
	\end{figure}

\clearpage
\bibliography{references}

\begin{thebibliography}{10}

\bibitem{moll14}
Yves Achdou, Francisco~J. Buera, Jean-Michel Lasry, Pierre-Louis Lions, and
  Benjamin Moll.
\newblock Partial differential equation models in macroeconomics.
\newblock {\em Philos. Trans. R. Soc. Lond. Ser. A Math. Phys. Eng. Sci.},
  372(2028):20130397, 19, 2014.

\bibitem{achdou2010mean}
Yves Achdou and Italo Capuzzo-Dolcetta.
\newblock Mean field games: numerical methods.
\newblock {\em SIAM Journal on Numerical Analysis}, 48(3):1136--1162, 2010.

\bibitem{moll17}
Yves Achdou, Jiequn Han, Jean-Michel Lasry, Pierre-Louis Lions, and Benjamin
  Moll.
\newblock Income and wealth distribution in macroeconomics: A continuous-time
  approach.
\newblock Working Paper 23732, National Bureau of Economic Research, August
  2017.

\bibitem{anil2019sorting}
Cem Anil, James Lucas, and Roger Grosse.
\newblock Sorting out lipschitz function approximation.
\newblock In {\em International Conference on Machine Learning}, pages
  291--301, 2019.

\bibitem{arjovsky2017wasserstein}
Martin Arjovsky, Soumith Chintala, and L{\'e}on Bottou.
\newblock Wasserstein generative adversarial networks.
\newblock In {\em International Conference on Machine Learning}, pages
  214--223, 2017.

\bibitem{bellman1966dynamic}
Richard Bellman.
\newblock Dynamic programming.
\newblock {\em Science}, 153(3731):34--37, 1966.

\bibitem{benamou2017variational}
Jean-David Benamou, Guillaume Carlier, and Filippo Santambrogio.
\newblock Variational mean field games.
\newblock In {\em Active Particles, Volume 1}, pages 141--171. Springer, 2017.

\bibitem{cao2020approximation}
Haoyang Cao and Xin Guo.
\newblock Approximation and convergence of gans training: an sde approach.
\newblock {\em arXiv preprint arXiv:2006.02047}, 2020.

\bibitem{cao2020connecting}
Haoyang Cao, Xin Guo, and Mathieu Lauri{\`e}re.
\newblock Connecting gans and mfgs.
\newblock {\em arXiv:2002.04112}, 2020.

\bibitem{cardaliaguet2015mean}
Pierre Cardaliaguet and P~Jameson Graber.
\newblock Mean field games systems of first order.
\newblock {\em ESAIM: Control, Optimisation and Calculus of Variations},
  21(3):690--722, 2015.

\bibitem{cardaliaguet2015second}
Pierre Cardaliaguet, P~Jameson Graber, Alessio Porretta, and Daniela Tonon.
\newblock Second order mean field games with degenerate diffusion and local
  coupling.
\newblock {\em Nonlinear Differential Equations and Applications NoDEA},
  22(5):1287--1317, 2015.

\bibitem{cardialiaguet2018}
Pierre Cardaliaguet and Charles-Albert Lehalle.
\newblock Mean field game of controls and an application to trade crowding.
\newblock {\em Math. Financ. Econ.}, 12(3):335--363, 2018.

\bibitem{carmona2019linear}
Ren{\'e} Carmona, Mathieu Lauri{\`e}re, and Zongjun Tan.
\newblock Linear-quadratic mean-field reinforcement learning: convergence of
  policy gradient methods.
\newblock {\em arXiv:1910.04295}, 2019.

\bibitem{chu2020stability}
Kenji~Fukumizu Casey~Chu, Kentaro~Minami.
\newblock Smoothness and stability in gans.
\newblock {\em arXiv:2002.04185}, 2020.

\bibitem{jaimungal19}
Philippe Casgrain and Sebastian Jaimungal.
\newblock Algorithmic trading in competitive markets with mean field games.
\newblock {\em SIAM News}, 52(2), 2019.

\bibitem{chang2020game}
Sheryl~L Chang, Mahendra Piraveenan, Philippa Pattison, and Mikhail Prokopenko.
\newblock Game theoretic modelling of infectious disease dynamics and
  intervention methods: a review.
\newblock {\em Journal of Biological Dynamics}, 14(1):57--89, 2020.

\bibitem{chen2018neural}
Tian~Qi Chen, Yulia Rubanova, Jesse Bettencourt, and David~K Duvenaud.
\newblock Neural ordinary differential equations.
\newblock In {\em Advances in Neural Information Processing Systems}, pages
  6571--6583, 2018.

\bibitem{chow2017algorithm}
Yat~Tin Chow, J{\'e}r{\^o}me Darbon, Stanley Osher, and Wotao Yin.
\newblock Algorithm for overcoming the curse of dimensionality for
  time-dependent non-convex hamilton--jacobi equations arising from optimal
  control and differential games problems.
\newblock {\em Journal of Scientific Computing}, 73(2-3):617--643, 2017.

\bibitem{chow2018algorithm}
Yat~Tin Chow, J{\'e}r{\^o}me Darbon, Stanley Osher, and Wotao Yin.
\newblock Algorithm for overcoming the curse of dimensionality for certain
  non-convex hamilton--jacobi equations, projections and differential games.
\newblock {\em Annals of Mathematical Sciences and Applications},
  3(2):369--403, 2018.

\bibitem{chow2019algorithm}
Yat~Tin Chow, Wuchen Li, Stanley Osher, and Wotao Yin.
\newblock Algorithm for hamilton--jacobi equations in density space via a
  generalized hopf formula.
\newblock {\em Journal of Scientific Computing}, 80(2):1195--1239, 2019.

\bibitem{nurbekyan18}
Marco Cirant and Levon Nurbekyan.
\newblock The variational structure and time-periodic solutions for mean-field
  games systems.
\newblock {\em Minimax Theory Appl.}, 3(2):227--260, 2018.

\bibitem{darbon2016algorithms}
J{\'e}r{\^o}me Darbon and Stanley Osher.
\newblock Algorithms for overcoming the curse of dimensionality for certain
  hamilton--jacobi equations arising in control theory and elsewhere.
\newblock {\em Research in the Mathematical Sciences}, 3(1):19, 2016.

\bibitem{paola19}
A.~{De Paola}, V.~{Trovato}, D.~{Angeli}, and G.~{Strbac}.
\newblock A mean field game approach for distributed control of thermostatic
  loads acting in simultaneous energy-frequency response markets.
\newblock {\em IEEE Transactions on Smart Grid}, 10(6):5987--5999, Nov 2019.

\bibitem{dukler2019wasserstein}
Yonatan Dukler, Wuchen Li, Alex Lin, and Guido Montufar.
\newblock Wasserstein of wasserstein loss for learning generative models.
\newblock In {\em International Conference on Machine Learning}, pages
  1716--1725, 2019.

\bibitem{elamvazhuthi2019mean}
Karthik Elamvazhuthi and Spring Berman.
\newblock Mean-field models in swarm robotics: a survey.
\newblock {\em Bioinspiration \& Biomimetics}, 15(1):015001, 2019.

\bibitem{finlay2020howtotrain}
Chris Finlay, Bj\"orn-Henrik Jacobsen, Levon Nurbekyan, and Adam~M Oberman.
\newblock How to train your neural ode.
\newblock {\em arXiv:2002.02798}, 2020.

\bibitem{caines17}
D.~{Firoozi} and P.~E. {Caines}.
\newblock An optimal execution problem in finance targeting the market trading
  speed: An mfg formulation.
\newblock In {\em 2017 IEEE 56th Annual Conference on Decision and Control
  (CDC)}, pages 7--14, Dec 2017.

\bibitem{Fu2020Actor-Critic}
Zuyue Fu, Zhuoran Yang, Yongxin Chen, and Zhaoran Wang.
\newblock Actor-critic provably finds nash equilibria of linear-quadratic
  mean-field games.
\newblock In {\em International Conference on Learning Representations}, 2020.

\bibitem{pmlr-v84-genevay18a}
Aude Genevay, Gabriel Peyre, and Marco Cuturi.
\newblock Learning generative models with sinkhorn divergences.
\newblock In {\em Proceedings of the Twenty-First International Conference on
  Artificial Intelligence and Statistics}, volume~84 of {\em Proceedings of
  Machine Learning Research}, pages 1608--1617. PMLR, 2018.

\bibitem{gomes2015economic}
Diogo~A Gomes, Levon Nurbekyan, and Edgard~A Pimentel.
\newblock {\em Economic models and mean-field games theory}.
\newblock IMPA Mathematical Publications. Instituto Nacional de Matem\'atica
  Pura e Aplicada (IMPA), Rio de Janeiro, 2015.

\bibitem{gomes2018electricity}
Diogo~A. Gomes and J.~Sa\'{u}de.
\newblock A mean-field game approach to price formation in electricity markets.
\newblock {\em arXiv:1807.07088}, 2018.

\bibitem{goodfellow2014generative}
Ian Goodfellow, Jean Pouget-Abadie, Mehdi Mirza, Bing Xu, David Warde-Farley,
  Sherjil Ozair, Aaron Courville, and Yoshua Bengio.
\newblock Generative adversarial nets.
\newblock In {\em Advances in neural information processing systems}, pages
  2672--2680, 2014.

\bibitem{grathwohl2018ffjord}
Will Grathwohl, Ricky~TQ Chen, Jesse Betterncourt, Ilya Sutskever, and David
  Duvenaud.
\newblock {FFJORD}: Free-form continuous dynamics for scalable reversible
  generative models.
\newblock {\em International Conference on Learning Representations (ICLR)},
  2019.

\bibitem{gueant2011mean}
Olivier Gu{\'e}ant, Jean-Michel Lasry, and Pierre-Louis Lions.
\newblock Mean field games and applications.
\newblock In {\em Paris-Princeton lectures on mathematical finance 2010}, pages
  205--266. Springer, 2011.

\bibitem{gulrajani2017improved}
Ishaan Gulrajani, Faruk Ahmed, Martin Arjovsky, Vincent Dumoulin, and Aaron~C
  Courville.
\newblock Improved training of wasserstein gans.
\newblock In {\em Advances in neural information processing systems}, pages
  5767--5777, 2017.

\bibitem{guo2019learning}
Xin Guo, Anran Hu, Renyuan Xu, and Junzi Zhang.
\newblock Learning mean-field games.
\newblock In {\em Advances in Neural Information Processing Systems}, pages
  4967--4977, 2019.

\bibitem{guo2020general}
Xin Guo, Anran Hu, Renyuan Xu, and Junzi Zhang.
\newblock A general framework for learning mean-field games.
\newblock {\em arXiv preprint arXiv:2003.06069}, 2020.

\bibitem{haber2017stable}
Eldad Haber and Lars Ruthotto.
\newblock Stable architectures for deep neural networks.
\newblock {\em Inverse Problems}, 34(1):014004, 2017.

\bibitem{jacobs2019solving}
Matt Jacobs, Flavien L{\'e}ger, Wuchen Li, and Stanley Osher.
\newblock Solving large-scale optimization problems with a convergence rate
  independent of grid size.
\newblock {\em SIAM Journal on Numerical Analysis}, 57(3):1100--1123, 2019.

\bibitem{kizikale19}
Arman~C. Kizilkale, Rabih Salhab, and Roland~P. Malhamé.
\newblock An integral control formulation of mean field game based large scale
  coordination of loads in smart grids.
\newblock {\em Automatica}, 100:312 -- 322, 2019.

\bibitem{lasry2006jeux}
Jean-Michel Lasry and Pierre-Louis Lions.
\newblock Jeux {\`a} champ moyen. ii--horizon fini et contr{\^o}le optimal.
\newblock {\em Comptes Rendus Math{\'e}matique}, 343(10):679--684, 2006.

\bibitem{LasryLions2007}
Jean-Michel Lasry and Pierre-Louis Lions.
\newblock Mean field games.
\newblock {\em Jpn. J. Math.}, 2(1):229--260, 2007.

\bibitem{lee2020controlling}
Wonjun Lee, Siting Liu, Hamidou Tembine, and Stanley Osher.
\newblock Controlling propagation of epidemics via mean-field games.
\newblock {\em UCLA CAM preprint:20-19}, 2020.

\bibitem{li2016mean}
Sisi Li, Shengbo~Eben Li, and Kun Deng.
\newblock Mean-field control for improving energy efficiency.
\newblock In {\em Automotive Air Conditioning}, pages 125--143. Springer, 2016.

\bibitem{lin2018splitting}
Alex~Tong Lin, Yat~Tin Chow, and Stanley~J Osher.
\newblock A splitting method for overcoming the curse of dimensionality in
  hamilton--jacobi equations arising from nonlinear optimal control and
  differential games with applications to trajectory generation.
\newblock {\em Communications in Mathematical Sciences}, 16(7), 2018.

\bibitem{lin2018wasserstein}
Alex~Tong Lin, Wuchen Li, Stanley Osher, and Guido Mont{\'u}far.
\newblock Wasserstein proximal of gans.
\newblock {\em UCLA CAM preprint:18-53}, 2018.

\bibitem{lin2019fluid}
Jingrong Lin, Keegan Lensink, and Eldad Haber.
\newblock Fluid flow mass transport for generative networks.
\newblock {\em arXiv:1910.01694}, pages 1--2, 2019.

\bibitem{liu2020splitting}
Siting Liu and Levon Nurbekyan.
\newblock Splitting methods for a class of non-potential mean field games.
\newblock {\em arXiv preprint arXiv:2007.00099}, 2020.

\bibitem{liu2018mean}
Zhiyu Liu, Bo~Wu, and Hai Lin.
\newblock A mean field game approach to swarming robots control.
\newblock In {\em 2018 Annual American Control Conference (ACC)}, pages
  4293--4298. IEEE, 2018.

\bibitem{onken2020discretizeoptimize}
Derek Onken and Lars Ruthotto.
\newblock Discretize-optimize vs. optimize-discretize for time-series
  regression and continuous normalizing flows.
\newblock {\em arXiv:2005.13420}, 2020.

\bibitem{onken2020otflow}
Derek Onken, Samy Wu~Fung, Xingjian Li, and Lars Ruthotto.
\newblock {OT-Flow}: Fast and accurate continuous normalizing flows via optimal
  transport.
\newblock {\em arXiv:2006.00104}, 2020.

\bibitem{parkinson2020model}
Christian Parkinson, David Arnold, Andrea~L Bertozzi, and Stanley Osher.
\newblock A model for optimal human navigation with stochastic effects.
\newblock {\em arXiv:2005.03615}, 2020.

\bibitem{parzen1962}
Emanuel Parzen.
\newblock On estimation of a probability density function and mode.
\newblock {\em Ann. Math. Statist.}, 33(3):1065--1076, 09 1962.

\bibitem{peyre2019computational}
Gabriel Peyr{\'e}, Marco Cuturi, et~al.
\newblock Computational optimal transport.
\newblock {\em Foundations and Trends in Machine Learning}, 11(5-6):355--607,
  2019.

\bibitem{rosenblatt1956}
Murray Rosenblatt.
\newblock Remarks on some nonparametric estimates of a density function.
\newblock {\em Ann. Math. Statist.}, 27(3):832--837, 09 1956.

\bibitem{ruthotto2019deep}
Lars Ruthotto and Eldad Haber.
\newblock Deep neural networks motivated by partial differential equations.
\newblock {\em Journal of Mathematical Imaging and Vision}, pages 1--13, 2019.

\bibitem{ruthotto2020machine}
Lars Ruthotto, Stanley~J Osher, Wuchen Li, Levon Nurbekyan, and Samy~Wu Fung.
\newblock A machine learning framework for solving high-dimensional mean field
  game and mean field control problems.
\newblock {\em Proceedings of the National Academy of Sciences},
  117(17):9183--9193, 2020.

\bibitem{salimans2018improving}
Tim Salimans, Han Zhang, Alec Radford, and Dimitris Metaxas.
\newblock Improving {GAN}s using optimal transport.
\newblock In {\em International Conference on Learning Representations}, 2018.

\bibitem{sanjabi2018convergence}
Maziar Sanjabi, Jimmy Ba, Meisam Razaviyayn, and Jason~D Lee.
\newblock On the convergence and robustness of training gans with regularized
  optimal transport.
\newblock In {\em Advances in Neural Information Processing Systems}, pages
  7091--7101, 2018.

\bibitem{scott2005multidimensional}
David~W Scott and Stephan~R Sain.
\newblock Multidimensional density estimation.
\newblock {\em Handbook of statistics}, 24:229--261, 2005.

\bibitem{tanaka2019discriminator}
Akinori Tanaka.
\newblock Discriminator optimal transport.
\newblock In {\em Advances in Neural Information Processing Systems}, pages
  6813--6823, 2019.

\bibitem{Villani2003}
C{\'e}dric Villani.
\newblock {\em {Topics in Optimal Transportation}}.
\newblock American Mathematical Soc., 2003.

\bibitem{wai2018multi}
Hoi-To Wai, Zhuoran Yang, Zhaoran Wang, and Mingyi Hong.
\newblock Multi-agent reinforcement learning via double averaging primal-dual
  optimization.
\newblock In {\em Advances in Neural Information Processing Systems}, pages
  9649--9660, 2018.

\bibitem{weinan2019mean}
E~Weinan, Jiequn Han, and Qianxiao Li.
\newblock A mean-field optimal control formulation of deep learning.
\newblock {\em Research in the Mathematical Sciences}, 6(1):10, 2019.

\bibitem{yang2018deep}
Jiachen Yang, Xiaojing Ye, Rakshit Trivedi, Huan Xu, and Hongyuan Zha.
\newblock Deep mean field games for learning optimal behavior policy of large
  populations.
\newblock In {\em International Conference on Learning Representations}, 2018.

\end{thebibliography}
\bibliographystyle{plain}
	
\end{document}